\documentclass{article}

\usepackage[preprint]{spconf}
\usepackage{array}
\usepackage{booktabs}
\usepackage{cite}
\usepackage{graphicx}
\usepackage{cite}
\usepackage{amssymb,amsmath}
\usepackage{framed,multirow}
\usepackage{latexsym}
\usepackage{array}
\usepackage{balance}
\usepackage{booktabs}
\usepackage{caption}
\usepackage{color}
\usepackage{graphicx}
\usepackage{multirow}
\usepackage{times}
\usepackage{url}
\usepackage{threeparttable}
\usepackage{float}
\usepackage{bbding}
\usepackage{multicol}

\usepackage{hyperref}

\hypersetup{hidelinks, colorlinks=true, allcolors=black, pdfstartview=Fit, breaklinks=true}

\makeatletter
\newif\if@restonecol
\makeatother

\usepackage[linesnumbered,ruled,vlined]{algorithm2e}
\usepackage{algpseudocode}
\usepackage{amsmath}

\begin{document}
\title{DCQA: Document-Level Chart Question Answering towards Complex Reasoning and Common-Sense Understanding}

\name{\small{Anran Wu$^{\star}$, Luwei Xiao$^{\star}$, Xingjiao Wu$^{\dagger}$, Shuwen Yang$^{\star}$, Junjie Xu$^{\star}$, Zisong Zhuang$^{\star}$, Nian Xie$^{\sharp}$, Cheng Jin$^{\dagger}$, Liang He$^{\star}$}
\thanks{Anran Wu and Luwei Xiao contributed equally to this work. Corresponding author: Xingjiao Wu (e-mail: xjwu\_cs@fudan.edu.cn).}}

\address{$^{\star}$ East China Normal University, Shanghai, China\\
$^{\dagger}$Fudan University, Shanghai, China\\
$^{\sharp}$Huawei Noah's Ark Lab, Shenzheng, China
\\ \
}



\UseRawInputEncoding
\maketitle

\begin{abstract}

Visually-situated languages such as charts and plots are omnipresent in real-world documents. These graphical depictions are human-readable and are often analyzed in visually-rich documents to address a variety of questions that necessitate complex reasoning and common-sense responses. Despite the growing number of datasets that aim to answer questions over charts, most only address this task in isolation, without considering the broader context of document-level question answering. Moreover, such datasets lack adequate common-sense reasoning information in their questions. In this work, we introduce a novel task named document-level chart question answering (DCQA). The goal of this task is to conduct document-level question answering, extracting charts or plots in the document via document layout analysis (DLA) first and subsequently performing chart question answering (CQA). The newly developed benchmark dataset comprises 50,010 synthetic documents integrating charts in a wide range of styles (6 styles in contrast to 3 for PlotQA and ChartQA) and includes 699,051 questions that demand a high degree of reasoning ability and common-sense understanding. Besides, we present the development of a potent question-answer generation engine that employs table data, a rich color set, and basic question templates to produce a vast array of reasoning question-answer pairs automatically. Based on DCQA, we devise an OCR-free transformer for document-level chart-oriented understanding, capable of DLA and answering complex reasoning and common-sense questions over charts in an OCR-free manner. Our DCQA dataset is expected to foster research on understanding visualizations in documents, especially for scenarios that require complex reasoning for charts in the visually-rich document. We implement and evaluate a set of baselines, and our proposed method achieves comparable results.

\end{abstract}


\section{Introduction}
\label{sec:intro}

\begin{figure*}[t]
  \centering
  \includegraphics[width=1\linewidth]{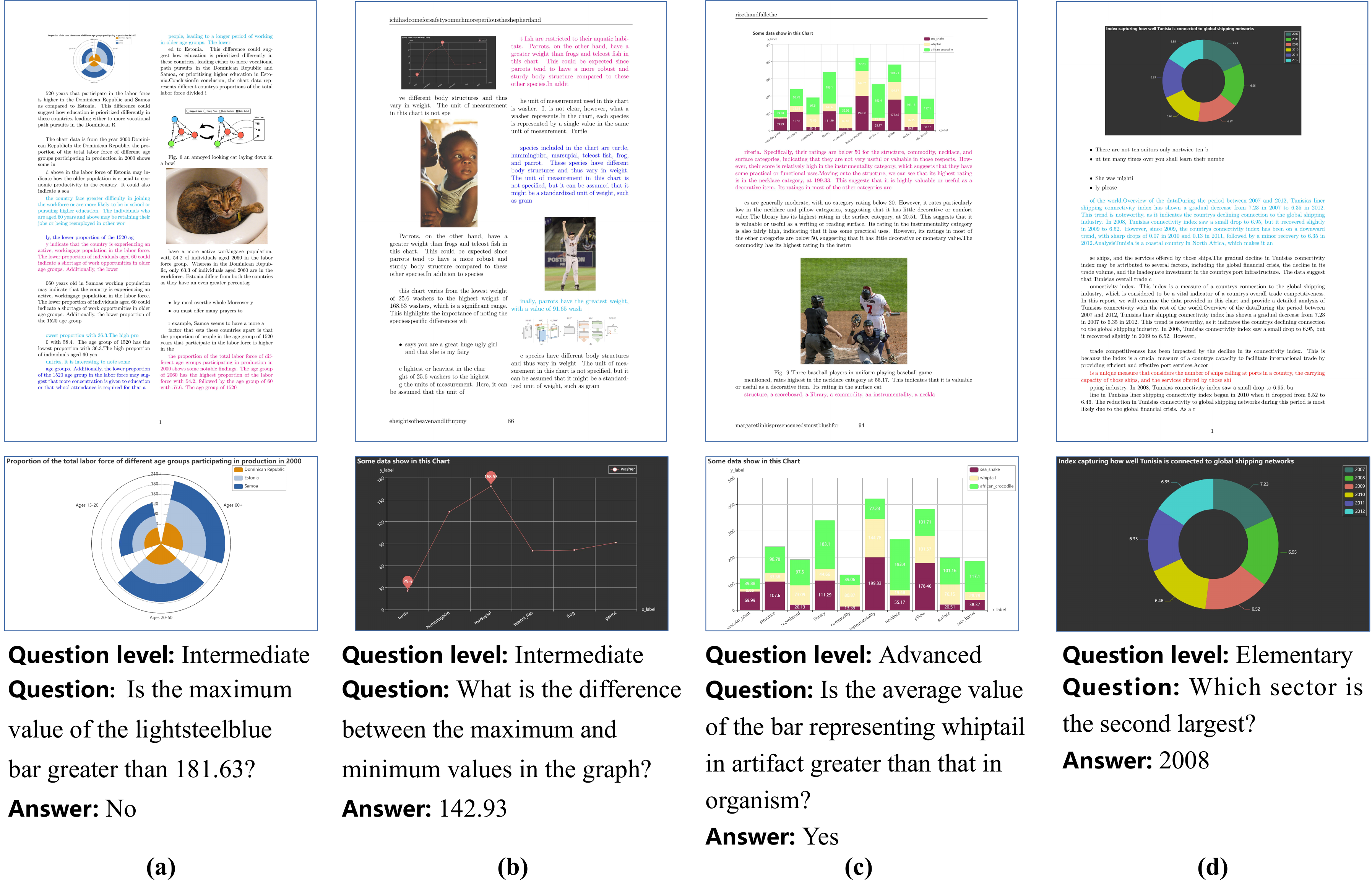}
  \caption{Examples from DCQA.}
  \label{example}
\end{figure*}

The emergence of visual language as a novel communicative tool, characterized by a tightly integrated interplay of visual and textual components, can be attributed to a confluence of factors, notably globalization, the growing intricacy of commerce and technology, and the convergence of lexicons from diverse fields that were once disparate~\cite{horn1998visual}. 
The prevalence of visually-situated language in various document types, such as academic, business, medical, and others, is markedly high~\cite{kim2020answering, lee2022pix2struct, ma2021towards}.
Gaining a comprehensive understanding of these graphical representations, such as charts and plots, is essential in extracting valuable and pragmatic insights from data~\cite{davila2020chart}. 
To conduct data analysis, individuals frequently pose intricate queries that require common-sense and arithmetic or logical operations pertaining to graphical representations. Answering such inquiries demands a substantial level of cognitive and reasoning exertion, as individuals are required to be aware of common sense and integrate numerous logical operations, including but not limited to retrieving entities, comparing trends, calculating averages, finding extremum, etc. Typically, the chart question answering (CQA) system~\cite{levy2022classification} aims to generate the desired answer by taking a chart-question pair as input, constituting a fundamental function within the domain of intelligent document understanding (IDU)~\cite{li2022dit}. 

Despite the CQA task has drawn ever-growing attention from visual question answering communities in recent years, existing datasets has encountered certain obstacles: \textit{(i)} Notably, while charts constitute crucial components of documents, the majority of current datasets treat the CQA task solely at the question-answering level, without taking into account its significance as a document-level task. \textit{(ii)} Questions generally prioritize reasoning or visual features, potentially losing sight of common sense information that individuals typically consider when posing questions, which is a misalignment with the typical questioning habits of individuals. \textit{(iii)} The quantity of chart types, as exemplified by PlotQA and ChartQA datasets, is comparatively restricted (only three). Such a limited representation fails to capture the broad range of chart styles that are present in real-world documents.
 
Furthermore, in real-world settings, users typically first identify the location of charts in documents before querying them. However, directly analyzing document layout poses challenges, as charts lack explicit annotations. Manually annotating datasets for document layout analysis related to charts is remarkably laborious and time-consuming. This motivates an automated system to generate annotations associated with charts, eliminating the need for costly labeled data collection. Such a system would locate charts in arbitrary documents without annotations and produce bounding boxes highlighting them, providing chart-specific layout information without human intervention. Additionally, certain baseline models rely on obtaining high-quality optical character recognition (OCR) outcomes to extract the data table structure from the chart image. Therefore, current models' efficacy generally relies upon the accuracy of OCR results. Nevertheless, incorporating an OCR-dependent approach for CQA system poses significant challenges. For one thing, commercially available OCR techniques often exhibit limited adaptability in addressing diverse languages or changes in the domain, which are commonly encountered in the context of charts. 
Such limitations may impede the generalization ability of these methods. 
For another, the occurrence of errors during the OCR process is unavoidable, and such erroneous outcomes have the potential to propagate to the CQA system, thereby adversely affecting subsequent processes~\cite{kim2022ocr}.

To alleviate above issues, we go beyond the traditional dataset by presenting a large-scale document-level chart question answering dataset (DCQA). DCQA comprises 50,010 synthetic documents and 699,051 question-answer pairs generated using our customized semantic-rich question-answer generation engine. The dataset includes questions that focus on vision, complex reasoning, and common-sense knowledge. Common-sense knowledge reasoning primarily involves evaluating the ability of CQA models to distinguish between legend labels and entity names belonging to specific parent classes, and subsequently performing reasoning operations based on this discriminative ability. Each document in the DCQA includes a chart, unrelated images and a descriptive caption related to the chart. The language used in the DCQA dataset is English. The chart types exhibit a diverse range of styles and can be broadly categorized into six major types, namely Bar chart, Line plot, Pie chart, Scatter plot, Box plot, and Mixed chart, each of which is further divided into subtypes, yielding a total of 30 chart subtypes. Figure~\ref{example} displays some examples from DCQA. More examples are provided in Appendix \ref{section:f}.

Drawing upon the DCQA dataset, we further devise a transformer-based  OCR-free architecture to perform document layout analysis and chart question answering. Initially, we exploit swin transformer~\cite{liu2021swin} as the vision backbone to extract visual features of the input document. Next, the extracted features are fed into the detection component to perform document layout analysis~\cite{binmakhashen2019document}. Upon successfully identifying the chart image, we extract the relevant visual content from the chart, which is then utilized as input to the textual decoder for answer prediction. This novel OCR-free architecture provides a plug-and-play solution for performing chart question answering directly from the document.

In a nutshell, our contributions are as follows:

\begin{itemize}
\item We present a comprehensive and extensive document-level chart question answering dataset, DCQA, which features a wide range of chart styles and includes question-answer pairs that incorporate complex reasoning and common-sense knowledge. The dataset's scale and diversity make it a valuable resource for researchers interested in developing and evaluating chart question answering models.
\item We conceptualize chart question answering as a document-level task and propose a transformer-based OCR-free model to effectively address this task.

\item We perform comprehensive experiments and thorough analyses on DCQA, verifying the efficacy of our model.
\end{itemize}

\begin{table*}[t]
  \centering
  \caption{A comprehensive comparison between existing datasets and our proposed DCQA.}
  \resizebox{1\textwidth}{!}{
  \begin{tabular}
  {@{}l|cccc|cccc|cl@{}}
  \toprule
  \multicolumn{1}{c|}{\multirow{2}{*}{Datasets}} & \multicolumn{4}{c|}{Chart}                                                                                                                                           & \multicolumn{4}{c|}{Question}                                                                                       & \multicolumn{2}{c}{\multirow{2}{*}{Available}} \\ \cmidrule(lr){2-9}
  \multicolumn{1}{c|}{}                          & \multicolumn{1}{l|}{Document-level} & \multicolumn{1}{c|}{Types} & \multicolumn{1}{c|}{Num.}    & Source                                                             & \multicolumn{1}{l|}{Com. sense} & \multicolumn{1}{c|}{Types}          & \multicolumn{1}{c|}{Templates} & Num.       & \multicolumn{2}{c}{}                           \\ \midrule
  FigureQA~\cite{kahou2017figureqa}                                       & \multicolumn{1}{c|}{\XSolidBrush}               & \multicolumn{1}{c|}{4}     & \multicolumn{1}{c|}{180,000} & Colour set                                                         & \multicolumn{1}{c|}{\XSolidBrush}           & \multicolumn{1}{c|}{Template based} & \multicolumn{1}{c|}{15}        & 2,388,698  & \multicolumn{2}{c}{\Checkmark}                          \\
  DVQA~\cite{kafle2018dvqa}                                           & \multicolumn{1}{c|}{\XSolidBrush}              & \multicolumn{1}{c|}{1}     & \multicolumn{1}{c|}{300,000} & Brown corpus                                                       & \multicolumn{1}{c|}{\XSolidBrush}            & \multicolumn{1}{c|}{Template based} & \multicolumn{1}{c|}{26}        & 3,487,194  & \multicolumn{2}{c}{\Checkmark}                          \\
  LEAF-QA~\cite{chaudhry2020leaf}                                        & \multicolumn{1}{c|}{\XSolidBrush}              & \multicolumn{1}{c|}{5}     & \multicolumn{1}{c|}{245,903} & Real-world                                                         & \multicolumn{1}{c|}{\XSolidBrush}            & \multicolumn{1}{c|}{Template based} & \multicolumn{1}{c|}{35}        & 1,906,486  & \multicolumn{2}{c}{\XSolidBrush}                          \\
  LEAF-QA++~\cite{singh2020stl}                                      & \multicolumn{1}{c|}{\XSolidBrush}               & \multicolumn{1}{c|}{5}     & \multicolumn{1}{c|}{245,903} & Real-world                                                         & \multicolumn{1}{c|}{\XSolidBrush}            & \multicolumn{1}{c|}{Template based} & \multicolumn{1}{c|}{75}        & 2,589,355  & \multicolumn{2}{c}{\XSolidBrush}                          \\
  PlotQA-D1~\cite{methani2020plotqa}                                      & \multicolumn{1}{c|}{\XSolidBrush}               & \multicolumn{1}{c|}{3}     & \multicolumn{1}{c|}{224,377} & Real-world                                                         & \multicolumn{1}{c|}{\XSolidBrush}            & \multicolumn{1}{c|}{Template based} & \multicolumn{1}{c|}{74}        & 8,190,674  & \multicolumn{2}{c}{\Checkmark}                            \\
  PlotQA-D2~\cite{methani2020plotqa}                                      & \multicolumn{1}{c|}{\XSolidBrush}               & \multicolumn{1}{c|}{3}     & \multicolumn{1}{c|}{224,377} & Real-world                                                         & \multicolumn{1}{c|}{\XSolidBrush}            & \multicolumn{1}{c|}{Template based} & \multicolumn{1}{c|}{74}        & 28,952,641 & \multicolumn{2}{c}{\Checkmark}                           \\
  ChartQA-H~\cite{masry2022chartqa}                                      & \multicolumn{1}{c|}{\XSolidBrush}              & \multicolumn{1}{c|}{3}     & \multicolumn{1}{c|}{4,804}   & Real-world                                                         & \multicolumn{1}{c|}{\XSolidBrush}           & \multicolumn{1}{c|}{Human authored} & \multicolumn{1}{c|}{None}      & 9,608      & \multicolumn{2}{c}{\Checkmark}                           \\ \midrule
  DCQA                                           & \multicolumn{1}{c|}{\Checkmark}              & \multicolumn{1}{c|}{6}     & \multicolumn{1}{c|}{50,010}  & \begin{tabular}[c]{@{}c@{}}Real-world \& \\ WordNet\end{tabular} & \multicolumn{1}{c|}{\Checkmark}           & \multicolumn{1}{c|}{Template based} & \multicolumn{1}{c|}{324}       & 699,051     & \multicolumn{2}{c}{\Checkmark}                           \\ \bottomrule
  \end{tabular}
  }
  
  \label{tab:Comparison}
  \end{table*}

\section{Related work}
\label{sec:related}

\subsection{Chart Question Answering}
Given the critical role of the CQA subtask in Document AI and the widespread usage of OCR in CQA baseline systems for utilizing layout information, this paper proposes categorizing existing frameworks into two categories: OCR-free and OCR-based. For OCR-free methods, a series of previous studies~\cite{kahou2017figureqa, reddy2019figurenet, zou2020affinity} utilized a combination of Convolutional Neural Networks (CNN) and Long Short-term Memory (LSTM) units to respectively model the visual and textual representations. These modalities were then inputted into a feed-forward network or a Relation Network~\cite{santoro2017simple} to predict the answer. Regarding OCR-based approaches, early works~\cite{chaudhry2020leaf, singh2020stl, kafle2018dvqa, kafle2020answering} applied visual encoders to represent the visual content extracted from the chart image. These methods predominantly utilized OCR-based dynamic encoding techniques to model the question representation by incorporating positional information of the textual elements present in the chart. Subsequently, an attention mechanism was employed to interactively learn the features from the chart and the question, which were then exploited by a softmax classification layer. Recently, a few studies~\cite{methani2020plotqa, masry2022chartqa} proposed to use OCR to extract visual elements (e.g., legends, x-axis-label, y-axis-label) from the plot and utilize them to reconstruct the underlying data table. 
They then utilized table QA algorithms in conjunction with the visual backbone to address this task. Although the OCR-based approaches have demonstrated commendable performance, they are still vulnerable to the impact of OCR noise and entail significant computational complexity.

\subsection{Chart Question Answering Datasets}

To date, only a limited number of datasets have been explicitly designed for chart question answering. 
These datasets include FigureQA~\cite{kahou2017figureqa}, DVQA~\cite{kafle2018dvqa}, LEAF-QA~\cite{chaudhry2020leaf}, LEAFQA++~\cite{singh2020stl}, PlotQA~\cite{methani2020plotqa} and ChartQA~\cite{masry2022chartqa}. 
Despite consisting of a diverse set of synthetic charts, FigureQA suffers from a lack of specificity in terms of chart element labeling, utilizing only generic titles and color names.
Furthermore, the questions are limited to a few template-based formats with binary "yes/no" answers.
DVQA is limited to a single chart type, namely the Bar chart, and suffers from inadequate semantic relations between the textual elements. 
(e.g., bar and legend labels are randomly selected words) as well as restricted Y-axis value ranges. 
Numeric answers are primarily integers in both the train and test sets and share the same values. As with FigureQA, all bar plots in DVQA are artificially generated, and the questions are based on a small number of templates. 
Both LEAF-QA and its extended version, LEAFQA++, are not publicly available. Besides, they share a significant limitation: the absence of regression question-answering pairs. 
This is evident from the question templates described in their reference and the discrete answer set employed. 
Although PlotQA is currently the largest publicly available dataset for CQA, it is limited by imbalanced question distribution, as it is heavily weighted towards data-related questions and lacks an appropriate proportion of queries pertaining to the visual characteristics of chart elements, including color and shape. 
Regarding ChartQA, it is the pioneer dataset to compile real-world charts with a blend of human-created QA pairs and machine-generated QA pairs. However, despite its innovative contribution, ChartQA is characterized by a limited size and encompasses only three distinct plot types. Furthermore, the paucity of question-answer pairs (two) per chart undermines the potential for a comprehensive understanding of the underlying information conveyed by these visualizations. This work presents a novel and intricate CQA dataset, which diverges from prior datasets in several respects. 
Firstly, DCQA is introduced, which reformulates the CQA task by integrating document layout analysis and chart question answering. Secondly, in addition to visual and complex reasoning questions, DCQA incorporates common sense-aware questions. Last but not least, DCQA covers a broad range of chart types (6 main types and 30 sub-types) with a wealth of color types (595) and different backgrounds (white and black). 
A comprehensive comparison between existing datasets and DCQA is presented in Table~\ref{tab:Comparison}.

\begin{figure}[t]
  \centering
  \includegraphics[width=0.8\linewidth]{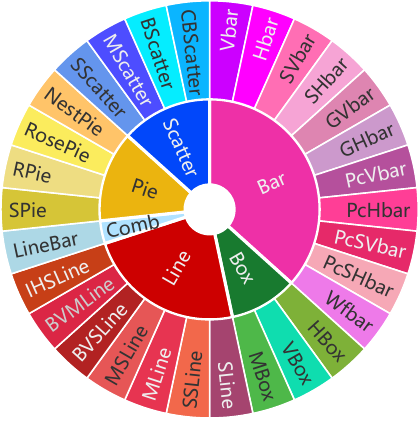}
  \caption{The illustration of chart types.}
  \label{chart_type}
\end{figure}

\begin{figure*}[t]
\centering
\includegraphics[width=1.0\linewidth]{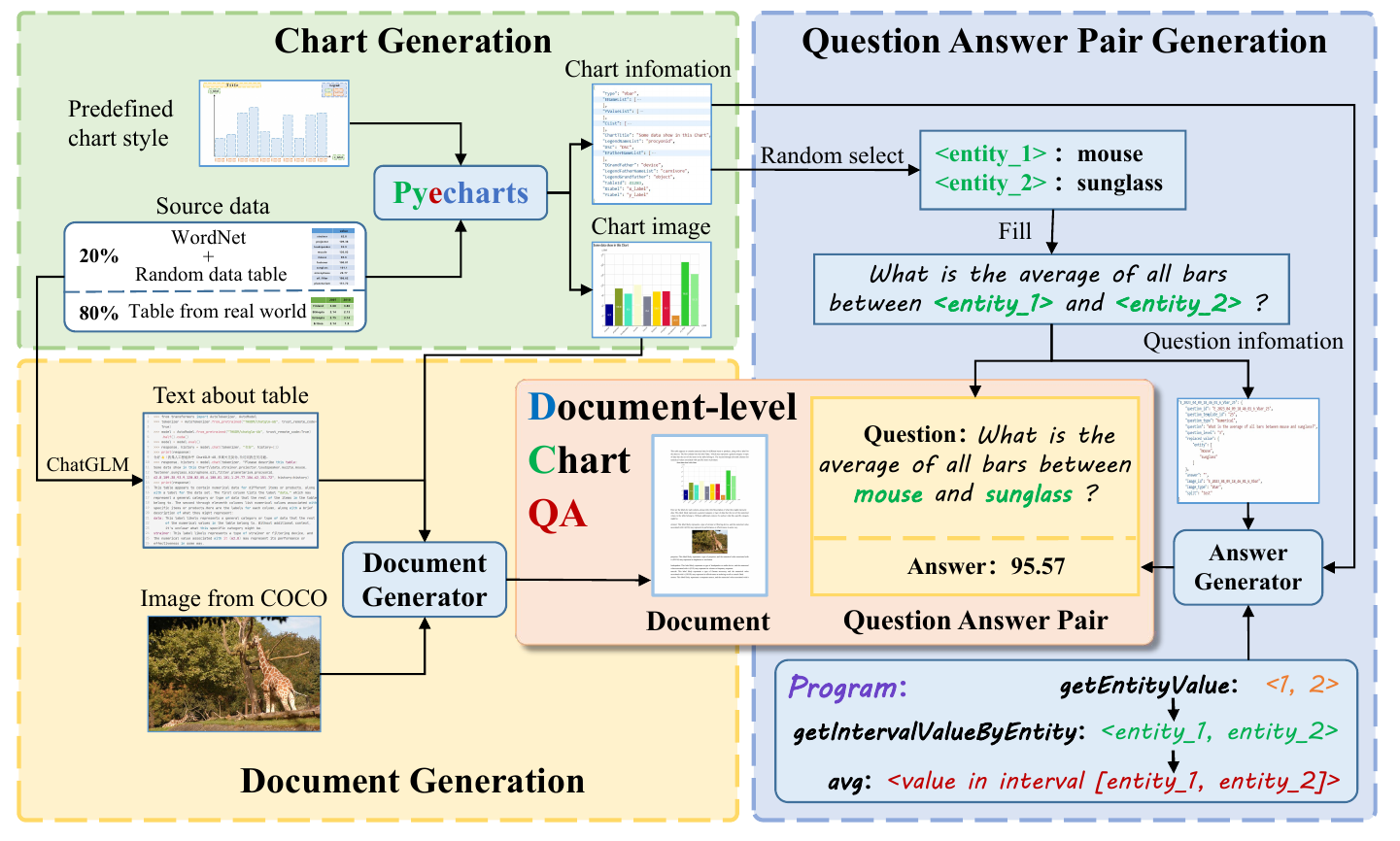}
\caption{The generation workflow of DCQA.}
\label{data_generate}
\end{figure*}

\section{DCQA Dataset}
\label{sec:method}

In this section, we describe the construction of DCQA from the following four aspects:\textit{(i)} Data collection, \textit{(ii)} Chart generation, \textit{(iii)} Automatic QA pair generation engine, and \textit{(iv)} Document generation. The general workflow of the DCQA generation process is shown in Figure~\ref{data_generate}. A detailed generation procedure is provided in Appendix \ref{section:b}.

\subsection{Data Collection}
Given the variability of chart styles in real-world scenarios, integrating real-world sources and randomly generated data for producing charts can augment the model’s robustness and adaptability to various chart formats encountered in practical scenarios. Drawing upon this observation, charts included in our dataset was derived by two means: utilizing real-world sources and randomly generated data. The detailed process of data collection is shown in Appendix \ref{section:b}.

\subsection{Chart Generation}
We exploit Pyecharts\footnote{https://github.com/pyecharts/pyecharts}, a Python visualization tool library based on the Echarts~\cite{li2018echarts} charting library, to generate our charts. Our DCQA contains six different chart styles: bar chart, line chart, scatter plot, box plot, pie chart, and combination chart (line and bar). These chart styles can be further divided into 30 sub-types (As shown in Figure~\ref{chart_type}). The color of chart elements is randomly picked up from a color set Johndecember, which covers a wide range of colors (595). Besides, the chart presents two distinct styles of background, namely dark and light, of which the former was not previously observed in any of the CQA datasets. 
Detailed chart information is provided in Appendix \ref{section:c}.

\subsection{Automatic QA Pair Generation Engine}
\subsubsection{Question Collection via Crowd-sourcing.}
Since the generated charts are from disparate data sources and encompass a wide range of topics, engaging a cadre of individuals with diverse backgrounds, experiences, and expertise is necessary to craft questions about the corresponding charts. We have meticulously curated a corpus of 573 charts spanning six categories, comprising data extracted from real-world and randomly generated sources, which serve as paradigmatic instances from which questions can be formulated. We commission a cohort of post-graduate students affiliated with our academic institution, and employees from Huawei company
, to generate ten distinct questions for each of the selected charts, with an emphasis on reasoning and common sense awareness. We have obtained 5730 questions.

\subsubsection{Question Generation.}
After an exhaustive process of meticulous meetings and in-depth discussions, we have successfully distilled a total of 324 question templates from the original pool of 5730 questions. Out of these templates, 204 are specifically tailored for visual and numeric reasoning, while 120 templates are dedicated to common sense reasoning. Code will be publicly available at 
\href{https://github.com/AnranWu-RichPo/DCQA}{github}
. Table~\ref{tab:dataset statistic} displays the statistics of the dataset. Details in Appendix \ref{section:d_2}.

\textbf{Visual and numeric reasoning:} These kinds of questions necessitate combing visual elements understanding and numerical reasoning techniques (e.g., sum, multiple, average, etc.). Integrating visual and numerical reasoning inquiries can facilitate CQA systems' comprehension of chart content, as it encourages the concurrent utilization of their visual and analytic faculties, thereby enabling them to engage in a more profound exploration of the underlying message conveyed by the data and achieve a more precise interpretation of chart figures. Examples of this type of question are presented in Figure~\ref{example} (a) and (b).

\begin{table*}[t]
\caption{Detailed statistics of DCQA.}
\centering
\resizebox{1\textwidth}{!}{\begin{tabular}{l|cccccc|cc|ccc} 
\toprule
\multirow{2}{*}{\begin{tabular}[c]{@{}c@{}}Dataset\\ split\end{tabular}} & \multicolumn{6}{c|}{Chart style}              & \multicolumn{2}{c|}{Question type} & \multicolumn{3}{c}{Answer type}  \\ 
\cmidrule(l){2-12}
 & Bar   & Line  & Pie  & Scatter & Box  & Comb. & Reasoning & Com. sense             & Yes/No & Elements & Open vocab.  \\ 
\midrule
Training                                                                 & 14,674 & 9,338  & 5,336 & 5,336    & 4,002 & 1,334  & 496,241    & 59,082                  & 171,447 & 51,215    & 332,661       \\
Validation                                                               & 1,826  & 1,162  & 664  & 664     & 498  & 166   & 61,752     & 9,058                   & 22,638  & 6,513     & 41,659        \\
Test                                                                     & 1,837  & 1,169  & 668  & 668     & 501  & 167   & 62,124     & 10,794                  & 23,261  & 6,688     & 42,969        \\ 
\midrule
Total                                                                    & 18,337 & 11,669 & 6,668 & 6,668    & 5,001 & 1,667  & 620,117    & 78,934                  & 217,346 & 64,416    & 417,289       \\
\bottomrule
\end{tabular}}
\label{tab:dataset statistic}
\end{table*}

\begin{table}
\caption{The distribution of different question levels.}
\centering
\resizebox{0.45\textwidth}{!}{\begin{tabular}{@{}l|ccccc@{}}
\toprule
                     \\ 
\multirow{2}{*}{Split} & \multicolumn{5}{c}{Question levels}                      \\ \cmidrule(l){2-6} 
                       & Beginner & Elementary & Intermediate & Advanced & Expert \\ \midrule
Train                  & 49,358    & 157,679    & 239,720      & 96,637    & 11,929   \\
Val.                   & 6,142     & 19,630      & 29,838        & 13,244    & 1,956    \\
Test                   & 6,179     & 19,746      & 30,053        & 14,071    & 2,869   \\ \midrule
Total                  & 61,679    & 197,055     & 299,611       & 123,952   & 16,754   \\ \bottomrule
\end{tabular}}

\label{tab:question-level}
\end{table}

\textbf{Common sense reasoning:} Questions of this type demand combining common sense knowledge and numerical operation. Common sense is able to serve as a facilitator for CQA systems to gain a more profound understanding of the real-life background and context reflected by the data, thus accurately inferring the meaning behind the data. Meanwhile, numerical reasoning skills can allow readers to fathom the underlying interconnections and relationships of the data and infer potential outcomes and trends. The combination of both proficiencies can profoundly equip CQA models with diverse conceptualizations of chart content and enable them to increase the usefulness of data in comprehensively examining and scrutinizing data, identifying patterns and trends, and making predictions and decisions. Examples of this type of question are presented in Figure~\ref{example} (c).

\textbf{Construction of the hierarchical entity database:} Common sense reasoning is a crucial aspect of DCQA, which primarily involves evaluating a CQA model's ability to discriminate between legend labels and entity names that belong to a specific parent class and then perform reasoning operations based on this discriminatory capacity. Therefore, a hierarchical entity database with a tree-structured architecture and a well-defined set of parent-child relationships is necessary to serve as a source for both entity names and legend labels. The construction of the hierarchical entity database is expounded upon in Appendix \ref{section:d_1}. 

\vspace{6pt}
\textbf{Categorization of question difficulty levels:}
Additionally, the entire set of question templates has been classified into five distinct levels delineated by their respective levels of complexity, namely, beginner, elementary, intermediate, advanced, and expert.

The statistic of question levels is displayed in 
Table~\ref{tab:question-level}.
The difficulty levels of the question templates are manually annotated based on the following criteria: (1) Beginner includes questions about the overall nature of a chart image, such as whether it is horizontal or vertical or the number of columns it contains, as well as retrieval for the value of a specific chart element. (2) Elementary primarily involves questions carrying out some form of operation on all chart elements within a chart, such as determining the maximum, minimum, median, or mean value. (3) Intermediate refers to questions that involve applying specific operations to chart elements that satisfy predetermined criteria, including but not limited to identifying the maximum, minimum, median, mean, sum, or difference of chart elements based on their color, legend, or numerical value. (4) Advanced questions demand performing composite operations on chart elements that meet a specific property (building upon the operations mentioned for intermediate questions), such as finding the sum or difference of two maximum values after they have been identified. (5) Building upon advanced questions, questions that involve common sense will be categorized as expert-level.

\begin{figure}[t]
  \centering
  \includegraphics[width=1\linewidth]{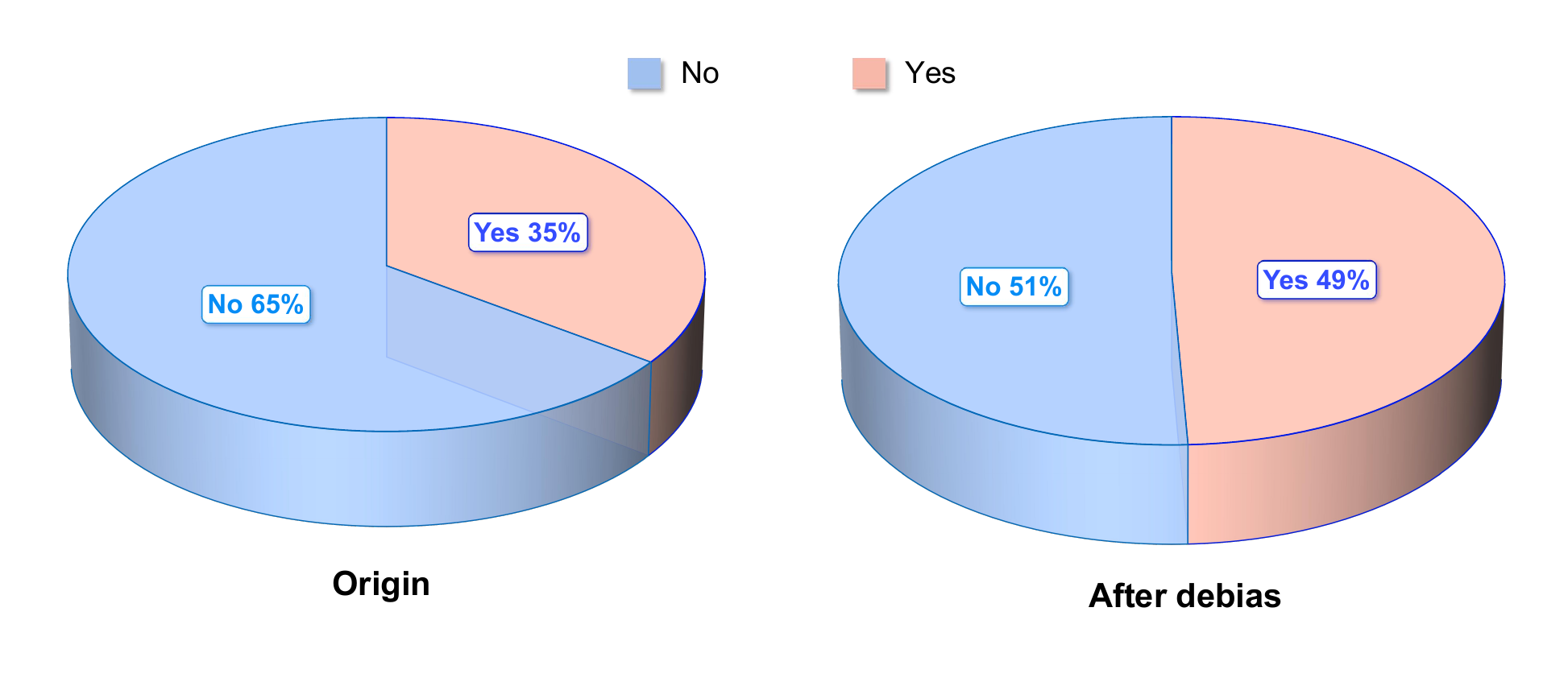}
  \caption{Comparison of Yes and No answer proportions for biased Yes/No questions before and after debiasing.}
  \label{debias_effect5}
\end{figure}

\begin{figure*}[t]
  \centering
  \includegraphics[width=1\linewidth]{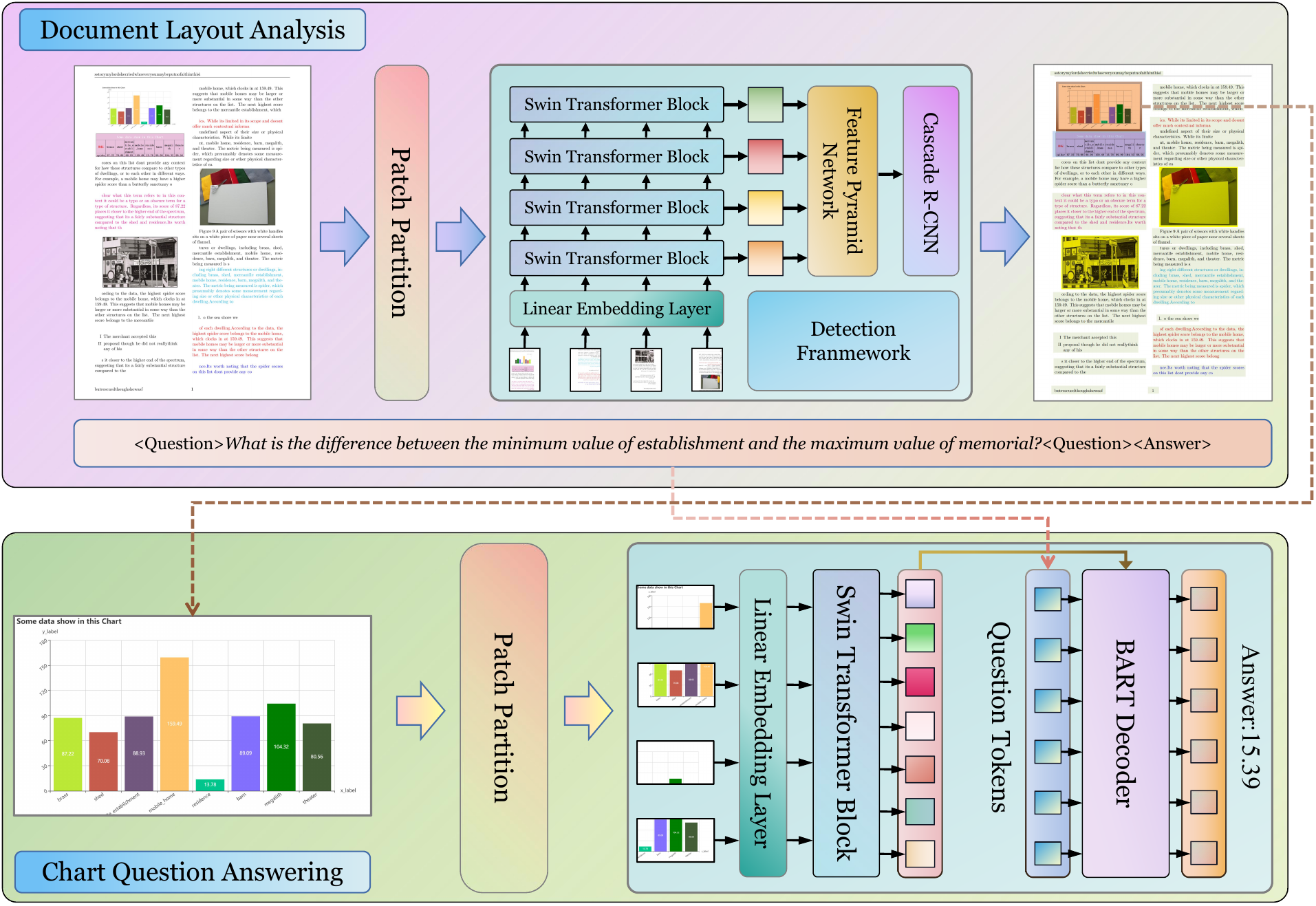}
  \caption{The overall architecture of the proposed TOT-Doctor.}
  \label{TAT-Doctor}
\end{figure*}

\subsubsection{Answer Generation.}
In this paper, answers are generated through an automated process based on a customized set of procedures. Specifically, a solution step is designed for each question template, with each step representing an atomic operation that is implemented using specific functions to achieve its intended functionality. When solving specific questions to generate answers, the designed solution steps are followed by invoking corresponding functions, resulting in answers for the respective questions. The process of obtaining answers and the advantages of automatic answer generation are elaborated upon in Appendix \ref{section:b}.

\subsubsection{Debiasing of Question-answer Pairs.}
Upon completing the question-answer pair generation process, extra analysis is conducted on the distribution of every answer type. 
It is noted that the highest proportion of the answer type in the dataset is the binary classification “yes” or “no”. However, the ratio between  “yes” and “no” is severely imbalanced, with a vastly larger number of “no” responses compared to “yes”. As a result, a post-processing adjustment is necessary to address this language bias and prevent the model from exploiting the answer distribution pattern to output the answer without paying attention to the visual content. 

The debiasing procedure can be concisely described as: Firstly, filter out all question templates with Yes/No answers. Secondly, count the number of Yes/No answers for each Yes/No question template in the dataset.
For each Yes/No question template, determine whether the number of Yes answers exceeds the number of No answers or vice versa. For the question with a larger proportion of answers, adjust the values of the replaceable modules in the question to change the answer to the less frequent one, and iterate this process until the number of Yes and No answers for the template are equal or differ by only 1.
Most of the Yes/No question templates can be balanced by modifying the answer using the above approach. However, a few questions cannot be balanced this way, and their answers are not significantly imbalanced. Therefore, we do not handle such question templates in this paper for now.
Before debiasing, the dataset's overall proportion of yes and no answers was 35.16$\%$ and 64.84$\%$, respectively. After debiasing, the overall proportion of yes and no answers in the dataset became 49.26$\%$ and 50.74$\%$, respectively. The effectiveness of the debiasing is compared in Figure~\ref{debias_effect5}, where the noticeable changes in the answer distribution of "Yes" and "No" before and after debiasing demonstrate that the debiasing method employed in this study effectively addresses the answer imbalance in binary questions.

\subsection{Document Generation} 
In accordance with the methodology outlined in~\cite{wu2021document}, we utilize LaTex to generate synthetic documents that include diverse multimedia elements. In addition to the chart image, the generated document also incorporates other visual content\footnote{https://cocodataset.org} and textual content produced by ChatGLM~\cite{du2022glm, zeng2022glm}. 
Notably, integrating these elements augments the informational value of the synthetic document beyond the mere inclusion of the chart image, which is more consistent with real-world documents. Detailed document information is provided
in Appendix \ref{section:e}.

\begin{table*}[t]
\centering
\caption{Hyperparameters of the baselines and the proposed method that were used on DCQA. The term "CE" refers to Cross Entropy.}
\renewcommand\arraystretch{1.3}

\resizebox{1\textwidth}{!}{
\begin{tabular}{lccccc}
\hline
Model         & LayoutLMv2 & LayoutLMv3 & Pix2struct      & Matcha          & Ours            \\ \hline
Training Loss & CE         & CE         & CE              & CE              & CE              \\
Batch Size    & 12         & 12         & 16              & 12              & 16              \\
Learning Rate & $10^{-5}$       & $2 \times 10^{-5}$       & $2 \times 10^{-5}$            & $2 \times 10^{-6}$            & $4 \times 10^{-5}$            \\
Optimizer     & AdamW      & AdamW      & AdamW           & AdamW           & AdamW           \\
Scheduler     & multi step & multi step & cosine schedule & cosine schedule & cosine schedule \\ \hline
\end{tabular}}

\label{hyper}
\end{table*}

\begin{table*}
\centering
\caption{Performance of the document layout analysis stage of ToT-Doctor on the DCQA test dataset (\%). PH stands for page header, PF stands for page footer, and PG represents page number.}
\renewcommand\arraystretch{1.3}
\resizebox{1\textwidth}{!}{
\begin{tabular}{llllllllll}
\hline
   & Chart imgae & Text   & Picture & Caption & List   & PH     & PF     & PG     & \textbf{Total}  \\ \hline
AP & 99.901      & 92.332 & 99.667  & 92.968  & 95.760 & 93.478 & 90.775 & 81.580 & \textbf{93.971} \\ \hline
\end{tabular}}

\label{tab:DLA-result}
\end{table*}

\begin{table}[t]
\caption{Performance of our TOT-Doctor and other baselines on DCQA. 
Best results are in bold. L2\_B indicates LayoutLMv2-base, L3\_B means LayoutLMv3-base.}

\begin{tabular}{@{}p{130pt}p{40pt}p{40pt}@{}}
\toprule
Method          & \multicolumn{1}{l}{Dev} & Test    \\ \midrule
L2\_B~\cite{xu2021layoutlmv2} & 4.844$\%$                & 4.816$\%$ \\
L3\_B~\cite{huang2022layoutlmv3} & 15.768$\%$                 & 15.760$\%$ \\
Pix2Struct-base~\cite{lee2022pix2struct} & 19.487$\%$                 & 19.294$\%$ \\
MATCHA-base~\cite{liu2022matcha}          & 19.249$\%$                 & 19.132$\%$ \\ \midrule
TOT-Doctor      & \textbf{40.658$\%$}                 & \textbf{40.090$\%$} \\ \bottomrule
\end{tabular}

\label{tab:main_results}
\end{table}

\begin{table*}[t]

\centering
\caption{Performance of our TOT-Doctor and other baselines on different question levels. Best results are in bold.}
\resizebox{1\textwidth}{!}
{\begin{tabular}{@{}lccccccccclcl@{}}
\toprule
& \multicolumn{2}{c}{LayoutLMv2} & \multicolumn{2}{c}{LayoutLMv3} & \multicolumn{2}{c}{Pix2Struct} & \multicolumn{2}{c}{MATCHA} & \multicolumn{4}{c}{TOT-Doctor}                                                                           \\ \cmidrule(l){2-13} 
\multirow{-2}{*}{Question levels} & Dev            & Test          & Dev            & Test          & Dev            & Test          & Dev          & Test        & \multicolumn{2}{c}{Dev}                                                    & \multicolumn{2}{c}{Test}    \\ \midrule
Beginner                          & 0.928$\%$        & 0.955$\%$       & 11.071$\%$        & 10.989$\%$       & 22.485$\%$        & 22.706$\%$       & 19.065$\%$      & 18.887$\%$     & \multicolumn{2}{c}{\textbf{44.302$\%$}} & \multicolumn{2}{c}{\textbf{42.337$\%$}} \\
Elementary                        & 3.316$\%$        & 3.150$\%$       & 14.544$\%$        & 14.671$\%$       & 18.945$\%$        & 18.373$\%$       & 17.835$\%$      & 17.371$\%$     & \multicolumn{2}{c}{\textbf{42.792$\%$}}                                                & \multicolumn{2}{c}{\textbf{41.948$\%$}}\\
Intermediate                      & 8.312$\%$        & 8.312$\%$       & 21.084$\%$        & 21.033$\%$       & 24.073$\%$        & 23.771$\%$       & 25.028$\%$      & 25.159$\%$     & \multicolumn{2}{c}{\textbf{47.543$\%$}}                                                & \multicolumn{2}{c}{\textbf{47.153$\%$}} \\
Advanced                          & 1.608$\%$        & 2.132$\%$       & 8.744$\%$        & 9.580$\%$       & 9.906$\%$        & 11.321$\%$       & 9.665$\%$      & 9.665$\%$     & \multicolumn{2}{c}{\textbf{24.268$\%$}}                                                & \multicolumn{2}{c}{\textbf{26.778$\%$}} \\
Expert                            & 1.483$\%$        & 1.150$\%$       & 9.254$\%$        & 8.609$\%$       & 10.429$\%$        & 10.491$\%$       & 10.429$\%$      & 10.422$\%$     & \multicolumn{2}{c}{\textbf{13.753$\%$}}                                                & \multicolumn{2}{c}{\textbf{13.768$\%$}} \\ \bottomrule
\end{tabular}}

\label{tab:question level}

\end{table*}

\begin{table*}[]

\centering
\caption{Performance of our TOT-Doctor and other baselines on different Answer types. Best results are in bold.}
\resizebox{1\textwidth}{!}
{\begin{tabular}{@{}lccccccccclcl@{}}
\toprule
                              & \multicolumn{2}{c}{LayoutLMv2} & \multicolumn{2}{c}{LayoutLMv3} & \multicolumn{2}{c}{Pix2Struct} & \multicolumn{2}{c}{MATCHA} & \multicolumn{4}{c}{TOT-Doctor}                                                                           \\ \cmidrule(l){2-13} 
\multirow{-2}{*}{Answer types} & Dev            & Test          & Dev            & Test          & Dev            & Test          & Dev          & Test        & \multicolumn{2}{c}{Dev}                                                    & \multicolumn{2}{c}{Test}    \\ \midrule
Yes/No                        & 15.152$\%$        & 15.098$\%$       & 48.944$\%$        & 48.996$\%$       & 49.249$\%$        & 49.142$\%$       & 50.362$\%$      & 50.209$\%$     & \multicolumn{2}{c}{\textbf{78.103$\%$}} & \multicolumn{2}{c}{\textbf{77.529$\%$}} \\
Numerical                     & 0$\%$            & 0$\%$           & 0.154$\%$        & 0.142$\%$       & 5.960$\%$        & 5.879$\%$       & 5.202$\%$      & 5.153$\%$     & \multicolumn{2}{c}{\textbf{24.650$\%$}}                                                & \multicolumn{2}{c}{\textbf{24.006$\%$}} \\
String                        & 0$\%$           & 0$\%$           & 0.322$\%$        & 0.508$\%$       & 2.564$\%$        & 1.675$\%$       & 0.952$\%$      & 0.867$\%$     & \multicolumn{2}{c}{\textbf{12.897$\%$}}                                                & \multicolumn{2}{c}{\textbf{13.218$\%$}} \\ \bottomrule
\end{tabular}}

\label{tab:answer types}
\end{table*}

\section{Method}
The DCQA dataset contains samples that more closely resemble real-world scenarios, considering the omnipresence of charts in documents. Our proposed approach for effectively comprehending chart-oriented documents is through the application of a novel OCR-free multi-modal CQA model, TOT-Doctor, \textbf{T}wo-stage \textbf{O}CR-free \textbf{t}ransformer for \textbf{doc}ument-level chart-\textbf{or}iented understanding. An overall architecture of the TOT-Doctor is displayed in Figure~\ref{TAT-Doctor}.

\subsection{Document Layout Analysis}

With the input document image denoted as $I \in \mathbb{R}^{3 \times H \times W}$, a patch partition module is utilized to divide it into non-overlapping patches. Subsequently, these patches are passed into a detection framework comprising several Swin Transformer-base blocks, a Feature Pyramid Network (FPN)~\cite{lin2017feature}, and a cascade R-CNN~\cite{cai2018cascade}. Initially, a linear embedding layer is employed to map the raw-valued RGB features of patches into a fixed-dimensional representation and then fed into Swin Transformer blocks. Since the scale of feature maps generated by different Swin Transformer blocks is distinct, they are fed into the multi-scale FPN. Next, the output from FPN is taken as the input for a Cascade R-CNN. Utilizing the prediction results of document objects, we extract the chart area and combine it with questions for chart understanding.

\subsection{Chart Question Answering}
After we obtain the chart image $I \in \mathbb{R}^{3 \times H_0 \times W_0}$ from the document via document layout analysis, at this stage, we utilize chart-question pairs to conduct chart question answering. Concretely, we model the visual features by dividing the chart image into non-overlapping patches via the patch partition module and passing through them into Swin Transformer-base. Following the teacher-forcing scheme~\cite{williams1989learning}, the chart visual features $V \in \mathbb{R}^{m \times d}$ obtained from the last Swin Transformer block is denoted as $K$ for BART decoder~\cite{liu2020multilingual}. In the training stage, the question representation $S \in \mathbb{R}^{n \times d}$ and the ground-truth $A \in \mathbb{R}^{k \times d}$ are concatenated as $Q$ for BART decoder. During the test phase, following the prompt scheme by GPT-3~\cite{brown2020language}, we add special tokens (as shown in Figure~\ref{TAT-Doctor}) to assist the model in generating the sequence. Subsequently, we feed them jointly into the BART decoder to generate the predicted sequence $\left(y_i\right)_{i=1}^z$, $y \in \mathbb{R}^{g}$ is the one-hot vector refers to the $i$-th generated token, $g$ denotes the size of the vocabulary, $z$ is a hyperparameter and indicates the number of tokens generated in the sequence.

\section{Experiments}

In this section, we provide a comprehensive analysis of the experimental results to establish the validity of the recently developed DCQA dataset and verify the excellent efficacy of our proposed TOT-Doctor model through a comparative evaluation against other baselines. 

\subsection{Baselines}
 We compare the TOT-Doctor with the classical approaches include LayoutLMv2~\cite{xu2021layoutlmv2}, LayoutLMv3~\cite{huang2022layoutlmv3}, Pix2Struct~\cite{lee2022pix2struct}, and MATCHA~\cite{liu2022matcha}. 
 \begin{itemize}
    \item LayoutLMv2~\cite{xu2021layoutlmv2} is a Transformer-based encoder that models multi-modal information, including text, layout, and image, by incorporating a spatial-aware self-attention mechanism along with two innovative pre-training strategies. 
    \item LayoutLMv3~\cite{huang2022layoutlmv3} is a multi-modal Transformer framework without the vision backbone that leverages reconstructive objectives for cross-modal alignment learning, showcasing notable generality in the context of document vision tasks.
    \item Pix2Struct~\cite{lee2022pix2struct} is an image-to-text model tailored for visual language understanding, which is pre-trained on visually-rich screenshots of web pages with \textit{screenshot parsing} objective.
    \item MATCHA~\cite{liu2022matcha} is a Pix2Struct-based model pre-trained for chart underlying structure understanding and mathematical reasoning.
\end{itemize}

\subsection{Evaluation Metrics}
Our study employs accuracy the primary evaluation metric, wherein the assessment of the predicted answer’s correctness depends on the nature of the answer type. In the case of textual answers, such as binary responses, entities, and integers, the evaluation criterion mandates that the predicted answer should match the ground truth exactly. For numerical answers in the form of floating-point values, it is not always feasible to expect that the predicted answer will precisely match the correct answer. Therefore, we consider the answer correct if it falls within 5$\%$ of the expected value.

\subsection{Experiment Setup}
This section primarily supplements experimental configurations, including parameters such as batch size and learning rate, and outlines the preprocessing steps undertaken to ensure a fair comparison for the extractive model.

Table \ref{hyper} shows the detailed experimental setup, in which CE refers to Cross Entropy. All models were verified every 5000 iterations during training. In our implementation of LayoutLMv2 and v3, we employ a multi-step learning rate schedule. More specifically, we gradually decrease the learning rate by a factor of 2 after each epoch of training. For other models, we use the cosine scheduler to adjust the learning rate, where the number of warm-up steps is set to 1000.
The LayoutLM series employs a model-based extraction approach, which requires the system to select answers from the optical character recognition (OCR) results. To enable the model to perform tasks such as binary classification (e.g., Yes/No), we have implemented a simple yet effective solution: we add two special characters, "Yes" and "No", to the OCR results.

\subsection{Model Parameter Quantity}
TOT-Doctor consists of two main components, namely the document layout analysis and the chart question answering. The model parameters of TOT-Doctor are calculated and found to be as follows: the encoder of the Swin Transformer used in the document layout analysis has 74M parameters, while the detector component has 48M parameters. The encoder of the Swin Transformer used in the chart question answering phase has 74M parameters, and the BART has 127M parameters.

\subsection{Main Results}

\subsubsection{Results of Document Layout Analysis}
Based on the fine-grained document element annotations present within the DCQA dataset introduced in this paper, encompassing various elements such as chart image, picture, textual content, list, caption, header, footer, and page number, the document layout analysis model of the proposed ToT-Doctor framework was trained. The conclusive results of the document layout analysis testing on the test set are presented in Table \ref{tab:DLA-result}. Notably, the detection accuracy for chart  image reached an impressive 99.901\%, exhibiting a near-complete precision in accurately identifying their respective spatial position. This achievement serves as a robust foundational prerequisite for facilitating subsequent stage of chart question answering task.

\subsubsection{Results of Question Answering}
The experiment results of our proposed TOT-Doctor and other baselines are displayed in Table~\ref{tab:main_results}. 
Due to the incapacity of the baseline to perform document layout analysis, we add our document layout analysis framework to them before conduct chart question answering.

Based on the data listed in Table~\ref{tab:main_results}, it can be seen that TOT-Doctor consistently surpasses its counterparts concerning the accuracy in both the validation and test sets, corroborating the efficacy of TOT-Doctor. It is noteworthy that despite LayoutLMv2 and LayoutLMv3 being reliant on OCR for obtaining the answers, their performance continues to lag behind our OCR-free TOT-Doctor. This observation proves the robustness and OCR error mitigation capabilities of the TOT-Doctor proposed in this study. Furthermore, in comparison to the latest state-of-the-art (SOTA) pre-trained visual language understanding model Pix2Struct, TOT-Doctor demonstrates superior performance, achieving a significant improvement of approximately 21.171$\%$ on the development dataset, and a respectable enhancement of 20.796$\%$ on the test dataset, respectively. The outstanding performance of our TOT-Doctor model underscores the significance of integrating vision and language features in an OCR-free manner to address the questions posed in DCQA effectively.

\subsection{Ablation Study}
\subsubsection{Evaluation on Different Question Levels.}
The DCQA dataset incorporates five distinct question levels. In order to better discern the effectiveness of our proposed TOT-Doctor and other baselines, we conduct a performance analysis on each question level. The results of our analysis are presented in Table~\ref{tab:question level} for reference. It is evident from the results that TOT-Doctor consistently outperforms other baselines in all five levels of questions. Notably, TOT-Doctor exhibits superior performance on intermediate-level questions, demonstrating its efficacy in directly applying specific operations such as identifying maximum, minimum, median, mean, and sum to chart elements or analyzing the differences of chart elements based on their color, legend, or numerical value. However, when encountering the sub-difficult advanced-level questions involving composite operations and the most challenging expert-level questions that necessitate commonsense understanding, the performance of all baselines, including TOT-Doctor, considerably decreases compared to the other three more tractable question levels. This implies that the ability for complex reasoning and commonsense understanding still requires further improvement for analyzing complex documents.

\begin{figure}[h]
  \centering

  \includegraphics[width=1\linewidth]{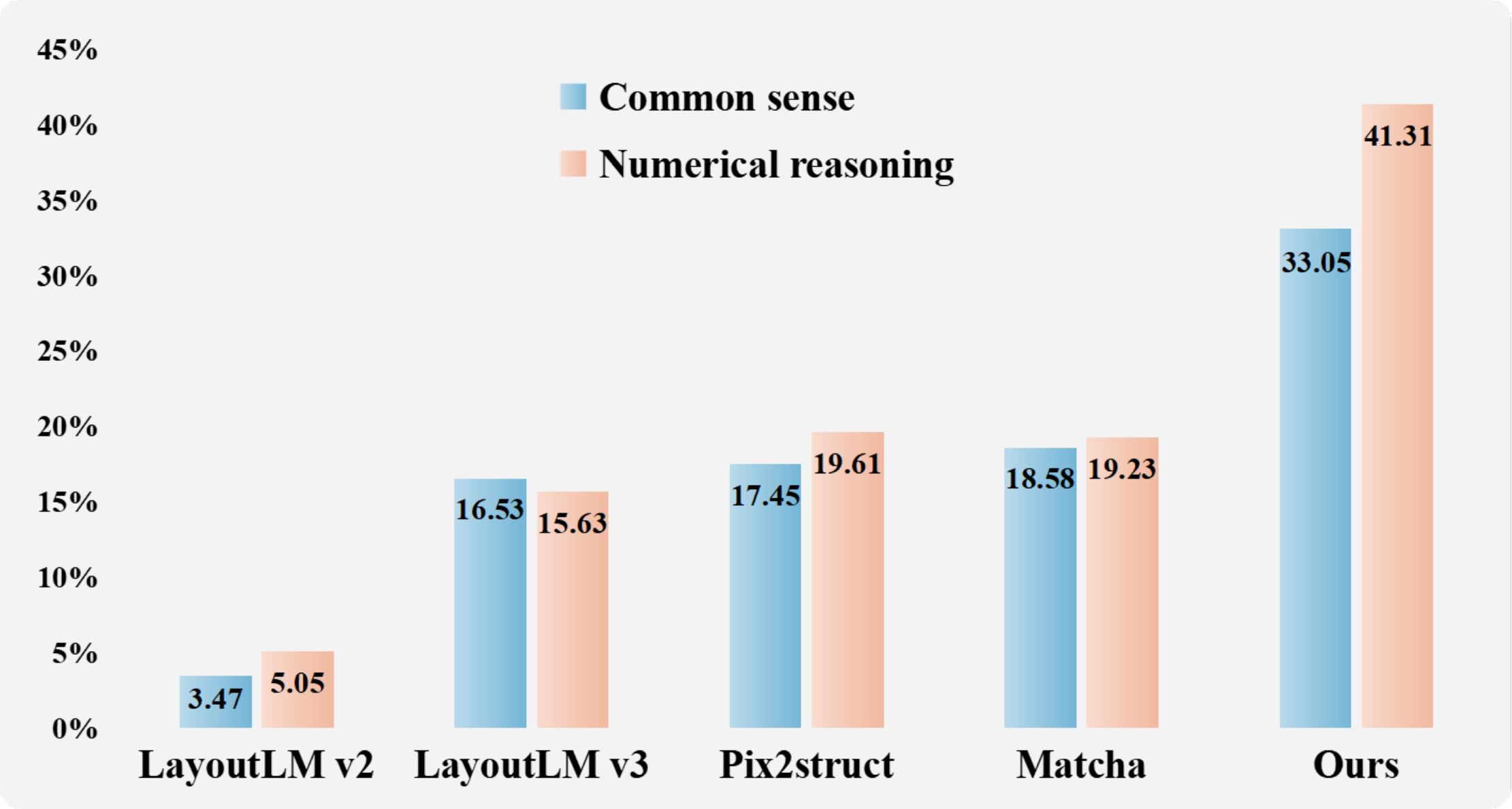}
  \caption{Performance comparison between common sense and numerical reasoning questions on DCQA test set.}

  \label{QuestionTypeAblation}
 
\end{figure}

\subsubsection{Evaluation on Different Question Types.}
We conduct additional experiments to evaluate the performance of the TOT-Doctor and baseline models on different question types. As discussed before, DCQA comprises two primary question types: visual and numeric reasoning and commonsense reasoning. Figure~\ref{QuestionTypeAblation} presents the results of all baselines for each question type on the test set. Our proposed TOT-Doctor outperforms other baselines significantly, particularly in numerical reasoning questions. To top it all off, TOT-Doctor demonstrates more proficiency in numerical reasoning compared to commonsense understanding.

\subsubsection{Evaluation on Different Answer Types.}

To further investigate our TOT-Doctor model, we assess the ability of the TOT-Doctor to generate different answer types. From Table~\ref{tab:answer types}, we discover that except for LayoutLMv2, all other baselines perform better in answering Yes/No questions. We observe that LayoutLMv2 and LayoutLMv3 exhibit frustrating performance in generating numerical and string answers. We speculate that this is mainly because LayoutLMv2 and LayoutLMv3 are extractive models, which means that they cannot generate answers that have not appeared in the document. This precisely explains their poor performance on the DCQA dataset, where the answers are largely obtained through data reasoning and involve numerical values. It is noted that EasyOCR\footnote{https://github.com/JaidedAI/EasyOCR} is utilized as the OCR system for these two baselines, which deviates from the OCR employed in their original versions. Based on this observation, we posit that the accuracy of the OCR system may have contributed to the subpar performance of the models. Moreover, TOT-Doctor is well versed in generating answers in terms of numerical or string, which verifies the robustness of TOT-Doctor. However, overall, all baselines achieve abysmal accuracy in generating numerical and string answers, highlighting the significant challenge posed by document-level chart understanding, which calls for further research efforts.

\section{Conclusion}
\label{sec:conclusion}

This paper introduces a newly developed and demanding dataset for document-level chart question answering called DCQA. The dataset comprises a sizeable body of documents containing a wide variety of chart styles. The questions posed within this dataset require complex reasoning skills and a strong capacity for common-sense understanding with regard to the information contained within the charts. We developed a specialized question-answer generation engine capable of automatically producing all question-answer pairs. Additionally, to tackle the highly challenging DCQA dataset, this paper proposes a novel OCR-free transformer-based framework, TOT-Doctor, designed explicitly for document-level chart question answering without relying on optical character recognition techniques. Experimental results demonstrate the veracity of our proposed TOT-Doctor in addressing complex document-level chart question answering tasks. However, the DCQA dataset has some limitations, such as the single background style of the charts, relatively simple document layouts, and the inability to exhaustively cover all possible questions generated by the templates. In future work, we plan to address these limitations by enriching the chart backgrounds with various colors and textures, considering more complex document layouts, and thereby enhancing the scalability of our method.

\small{
\bibliographystyle{IEEEbib}
\bibliography{total}

\begin{thebibliography}{10}

\bibitem{kahou2017figureqa}
Samira~Ebrahimi Kahou, Vincent Michalski, Adam Atkinson, {\'A}kos K{\'a}d{\'a}r, Adam Trischler, and Yoshua Bengio,
\newblock ``Figureqa: An annotated figure dataset for visual reasoning,''
\newblock {\em arXiv preprint arXiv:1710.07300}, 2017.

\bibitem{kafle2018dvqa}
Kushal Kafle, Brian Price, Scott Cohen, and Christopher Kanan,
\newblock ``Dvqa: Understanding data visualizations via question answering,''
\newblock in {\em Proceedings of the IEEE conference on computer vision and pattern recognition}, 2018, pp. 5648--5656.

\bibitem{chaudhry2020leaf}
Ritwick Chaudhry, Sumit Shekhar, Utkarsh Gupta, Pranav Maneriker, Prann Bansal, and Ajay Joshi,
\newblock ``Leaf-qa: Locate, encode \& attend for figure question answering,''
\newblock in {\em Proceedings of the IEEE/CVF Winter Conference on Applications of Computer Vision}, 2020, pp. 3512--3521.

\bibitem{singh2020stl}
Hrituraj Singh and Sumit Shekhar,
\newblock ``Stl-cqa: Structure-based transformers with localization and encoding for chart question answering,''
\newblock in {\em Proceedings of the 2020 Conference on Empirical Methods in Natural Language Processing (EMNLP)}, 2020, pp. 3275--3284.

\bibitem{methani2020plotqa}
Nitesh Methani, Pritha Ganguly, Mitesh~M Khapra, and Pratyush Kumar,
\newblock ``Plotqa: Reasoning over scientific plots,''
\newblock in {\em Proceedings of the IEEE/CVF Winter Conference on Applications of Computer Vision}, 2020, pp. 1527--1536.

\bibitem{masry2022chartqa}
Ahmed Masry, Do~Long, Jia~Qing Tan, Shafiq Joty, and Enamul Hoque,
\newblock ``Chartqa: A benchmark for question answering about charts with visual and logical reasoning,''
\newblock in {\em Findings of the Association for Computational Linguistics: ACL 2022}, 2022, pp. 2263--2279.

\bibitem{horn1998visual}
Robert~E Horn,
\newblock ``Visual language,''
\newblock {\em MacroVu Inc. Washington}, 1998.

\bibitem{kim2020answering}
Dae~Hyun Kim, Enamul Hoque, and Maneesh Agrawala,
\newblock ``Answering questions about charts and generating visual explanations,''
\newblock in {\em Proceedings of the 2020 CHI conference on human factors in computing systems}, 2020, pp. 1--13.

\bibitem{lee2022pix2struct}
Kenton Lee, Mandar Joshi, Iulia Turc, Hexiang Hu, Fangyu Liu, Julian Eisenschlos, Urvashi Khandelwal, Peter Shaw, Ming-Wei Chang, and Kristina Toutanova,
\newblock ``Pix2struct: Screenshot parsing as pretraining for visual language understanding,''
\newblock {\em arXiv preprint arXiv:2210.03347}, 2022.

\bibitem{ma2021towards}
Weihong Ma, Hesuo Zhang, Shuang Yan, Guangshun Yao, Yichao Huang, Hui Li, Yaqiang Wu, and Lianwen Jin,
\newblock ``Towards an efficient framework for data extraction from chart images,''
\newblock in {\em Document Analysis and Recognition--ICDAR 2021: 16th International Conference, Lausanne, Switzerland, September 5--10, 2021, Proceedings, Part I}. Springer, 2021, pp. 583--597.

\bibitem{davila2020chart}
Kenny Davila, Srirangaraj Setlur, David Doermann, Bhargava~Urala Kota, and Venu Govindaraju,
\newblock ``Chart mining: A survey of methods for automated chart analysis,''
\newblock {\em IEEE transactions on pattern analysis and machine intelligence}, vol. 43, no. 11, pp. 3799--3819, 2020.

\bibitem{levy2022classification}
Matan Levy, Rami Ben-Ari, and Dani Lischinski,
\newblock ``Classification-regression for chart comprehension,''
\newblock in {\em Computer Vision--ECCV 2022: 17th European Conference, Tel Aviv, Israel, October 23--27, 2022, Proceedings, Part XXXVI}. Springer, 2022, pp. 469--484.

\bibitem{li2022dit}
Junlong Li, Yiheng Xu, Tengchao Lv, Lei Cui, Cha Zhang, and Furu Wei,
\newblock ``Dit: Self-supervised pre-training for document image transformer,''
\newblock in {\em Proceedings of the 30th ACM International Conference on Multimedia}, 2022, pp. 3530--3539.

\bibitem{kim2022ocr}
Geewook Kim, Teakgyu Hong, Moonbin Yim, JeongYeon Nam, Jinyoung Park, Jinyeong Yim, Wonseok Hwang, Sangdoo Yun, Dongyoon Han, and Seunghyun Park,
\newblock ``Ocr-free document understanding transformer,''
\newblock in {\em Computer Vision--ECCV 2022: 17th European Conference, Tel Aviv, Israel, October 23--27, 2022, Proceedings, Part XXVIII}. Springer, 2022, pp. 498--517.

\bibitem{liu2021swin}
Ze~Liu, Yutong Lin, Yue Cao, Han Hu, Yixuan Wei, Zheng Zhang, Stephen Lin, and Baining Guo,
\newblock ``Swin transformer: Hierarchical vision transformer using shifted windows,''
\newblock in {\em Proceedings of the IEEE/CVF international conference on computer vision}, 2021, pp. 10012--10022.

\bibitem{binmakhashen2019document}
Galal~M Binmakhashen and Sabri~A Mahmoud,
\newblock ``Document layout analysis: a comprehensive survey,''
\newblock {\em ACM Computing Surveys (CSUR)}, vol. 52, no. 6, pp. 1--36, 2019.

\bibitem{reddy2019figurenet}
Revanth Reddy, Rahul Ramesh, Ameet Deshpande, and Mitesh~M Khapra,
\newblock ``Figurenet: A deep learning model for question-answering on scientific plots,''
\newblock in {\em 2019 International Joint Conference on Neural Networks (IJCNN)}. IEEE, 2019, pp. 1--8.

\bibitem{zou2020affinity}
Jialong Zou, Guoli Wu, Taofeng Xue, and Qingfeng Wu,
\newblock ``An affinity-driven relation network for figure question answering,''
\newblock in {\em 2020 IEEE International Conference on Multimedia and Expo (ICME)}. IEEE, 2020, pp. 1--6.

\bibitem{santoro2017simple}
Adam Santoro, David Raposo, David~G Barrett, Mateusz Malinowski, Razvan Pascanu, Peter Battaglia, and Timothy Lillicrap,
\newblock ``A simple neural network module for relational reasoning,''
\newblock {\em Advances in neural information processing systems}, vol. 30, 2017.

\bibitem{kafle2020answering}
Kushal Kafle, Robik Shrestha, Scott Cohen, Brian Price, and Christopher Kanan,
\newblock ``Answering questions about data visualizations using efficient bimodal fusion,''
\newblock in {\em Proceedings of the IEEE/CVF Winter conference on applications of computer vision}, 2020, pp. 1498--1507.

\bibitem{li2018echarts}
Deqing Li, Honghui Mei, Yi~Shen, Shuang Su, Wenli Zhang, Junting Wang, Ming Zu, and Wei Chen,
\newblock ``Echarts: a declarative framework for rapid construction of web-based visualization,''
\newblock {\em Visual Informatics}, vol. 2, no. 2, pp. 136--146, 2018.

\bibitem{wu2021document}
Xingjiao Wu, Yingbin Zheng, Tianlong Ma, Hao Ye, and Liang He,
\newblock ``Document image layout analysis via explicit edge embedding network,''
\newblock {\em Information Sciences}, vol. 577, pp. 436--448, 2021.

\bibitem{du2022glm}
Zhengxiao Du, Yujie Qian, Xiao Liu, Ming Ding, Jiezhong Qiu, Zhilin Yang, and Jie Tang,
\newblock ``Glm: General language model pretraining with autoregressive blank infilling,''
\newblock in {\em Proceedings of the 60th Annual Meeting of the Association for Computational Linguistics (Volume 1: Long Papers)}, 2022, pp. 320--335.

\bibitem{zeng2022glm}
Aohan Zeng, Xiao Liu, Zhengxiao Du, Zihan Wang, Hanyu Lai, Ming Ding, Zhuoyi Yang, Yifan Xu, Wendi Zheng, Xiao Xia, et~al.,
\newblock ``Glm-130b: An open bilingual pre-trained model,''
\newblock {\em arXiv preprint arXiv:2210.02414}, 2022.

\bibitem{xu2021layoutlmv2}
Yang Xu, Yiheng Xu, Tengchao Lv, Lei Cui, Furu Wei, Guoxin Wang, Yijuan Lu, Dinei Florencio, Cha Zhang, Wanxiang Che, et~al.,
\newblock ``Layoutlmv2: Multi-modal pre-training for visually-rich document understanding,''
\newblock in {\em Proceedings of the 59th Annual Meeting of the Association for Computational Linguistics and the 11th International Joint Conference on Natural Language Processing (Volume 1: Long Papers)}, 2021, pp. 2579--2591.

\bibitem{huang2022layoutlmv3}
Yupan Huang, Tengchao Lv, Lei Cui, Yutong Lu, and Furu Wei,
\newblock ``Layoutlmv3: Pre-training for document ai with unified text and image masking,''
\newblock in {\em Proceedings of the 30th ACM International Conference on Multimedia}, 2022, pp. 4083--4091.

\bibitem{liu2022matcha}
Fangyu Liu, Francesco Piccinno, Syrine Krichene, Chenxi Pang, Kenton Lee, Mandar Joshi, Yasemin Altun, Nigel Collier, and Julian~Martin Eisenschlos,
\newblock ``Matcha: Enhancing visual language pretraining with math reasoning and chart derendering,''
\newblock {\em arXiv preprint arXiv:2212.09662}, 2022.

\bibitem{lin2017feature}
Tsung-Yi Lin, Piotr Doll{\'a}r, Ross Girshick, Kaiming He, Bharath Hariharan, and Serge Belongie,
\newblock ``Feature pyramid networks for object detection,''
\newblock in {\em Proceedings of the IEEE conference on computer vision and pattern recognition}, 2017, pp. 2117--2125.

\bibitem{cai2018cascade}
Zhaowei Cai and Nuno Vasconcelos,
\newblock ``Cascade r-cnn: Delving into high quality object detection,''
\newblock in {\em Proceedings of the IEEE conference on computer vision and pattern recognition}, 2018, pp. 6154--6162.

\bibitem{williams1989learning}
Ronald~J Williams and David Zipser,
\newblock ``A learning algorithm for continually running fully recurrent neural networks,''
\newblock {\em Neural computation}, vol. 1, no. 2, pp. 270--280, 1989.

\bibitem{liu2020multilingual}
Yinhan Liu, Jiatao Gu, Naman Goyal, Xian Li, Sergey Edunov, Marjan Ghazvininejad, Mike Lewis, and Luke Zettlemoyer,
\newblock ``Multilingual denoising pre-training for neural machine translation,''
\newblock {\em Transactions of the Association for Computational Linguistics}, vol. 8, pp. 726--742, 2020.

\bibitem{brown2020language}
Tom Brown, Benjamin Mann, Nick Ryder, Melanie Subbiah, Jared~D Kaplan, Prafulla Dhariwal, Arvind Neelakantan, Pranav Shyam, Girish Sastry, Amanda Askell, et~al.,
\newblock ``Language models are few-shot learners,''
\newblock {\em Advances in neural information processing systems}, vol. 33, pp. 1877--1901, 2020.

\end{thebibliography}
}

\newpage
\clearpage

\appendix

\setcounter{figure}{0}
\renewcommand*{\thefigure}{A\arabic{figure}}

This supplementary material will provide more description, experiment and the details of DCQA dataset, 
which are organized as outlined below:

\begin{itemize}

\item Section \ref{section:b}: Dataset Generate Workflow.
This section provides further details of the DCQA generation workflow, supplemented by specific examples to facilitate a deeper comprehension.

\item Section \ref{section:c}: Chart Infomation.
This section presents the abundant chart information encompassed within the DCQA dataset, thereby furnishing researchers with a wealth of potential resources for future investigations.

\item Section \ref{section:d}: Question Generation.
This section delineates further intricacies in question generation within the DCQA dataset, alongside endeavors undertaken to enhance the robustness of question-answer pairs therein.

\item Section \ref{section:e}: Document Infomation.
This section elucidates the specifics of fine-grained annotation information within the DCQA dataset documents, which constitute a fundamental underpinning for the chart-oriented document layout analysis task.

\item Section \ref{section:f}: Document Example.
This section showcases examples of fine-grained categorization for chart types within the DCQA dataset, encompassing a total of 6 primary categories and 30 subcategories of charts.
\end{itemize}

\section{Dataset Generate Workflow}
\label{section:b}
\begin{figure*}[t]
  \centering
  \includegraphics[width=1\linewidth]{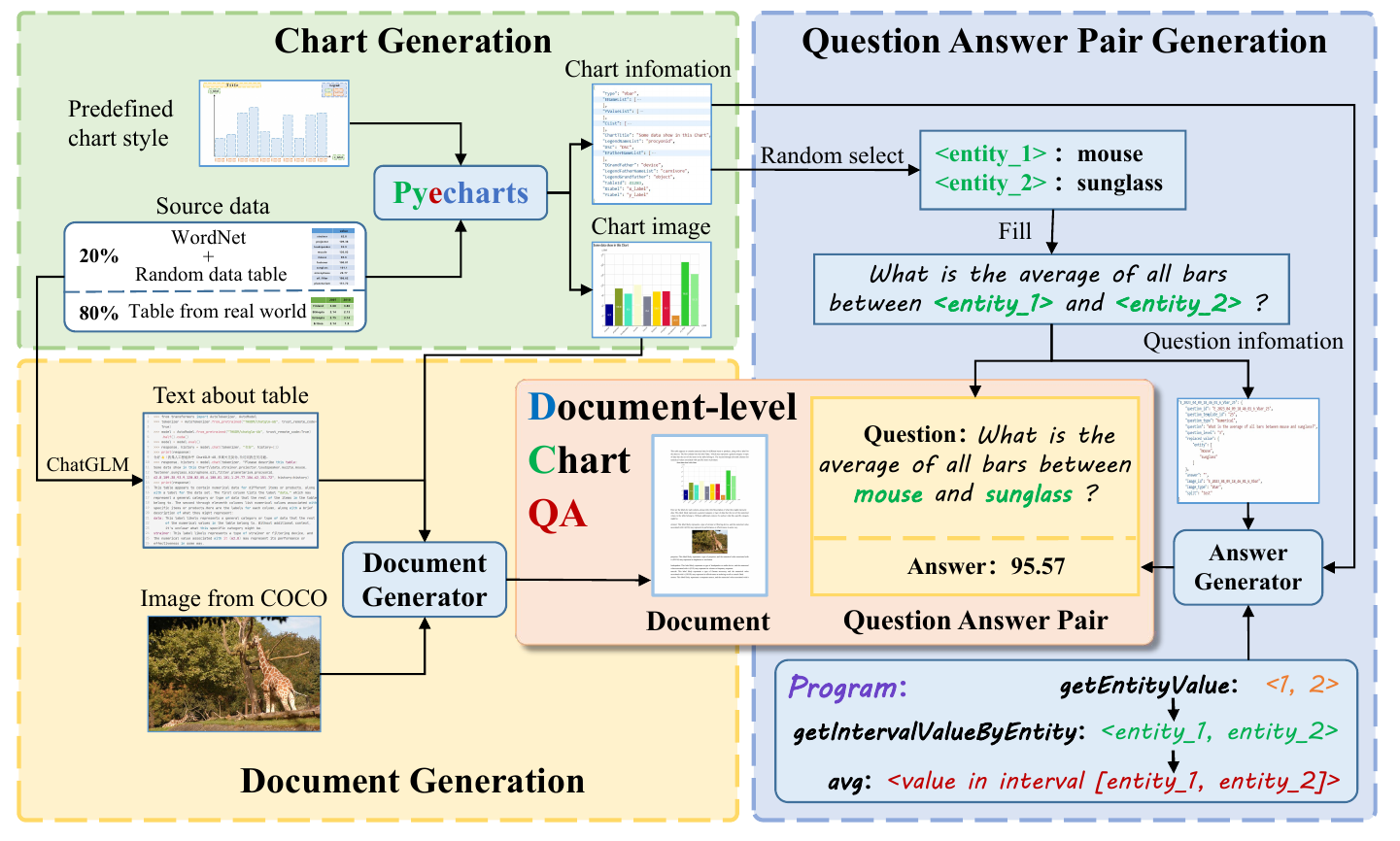}
  \caption{The generation workflow of DCQA. A larger version of Figure 2 is available in the body of the paper for a more detailed view of the process.}
  \label{data_generate1} 
\end{figure*}

The DCQA dataset generation process, as depicted in Figure~\ref{data_generate1}, encompasses four main components: chart generation, question generation, answer generation, and document generation. In addition, this section also encompasses the specifics of data collection.

\vspace{6pt}
\noindent

\textbf{Data Collection. } For the real-world data sources, we crawl data from the following on-line sources: (\textit{i}) World Bank Open Data (\url{https://data.worldbank.org}). (\textit{ii}) United Nations Data (\url{https://data.un.org/}). (\textit{iii}) International Monetary Fund Data (\url{https://data.imf.org/}). For the randomly generated data, we adopt part of the WordNet\footnote{https://wordnet.princeton.edu} as the chart element (e.g., legend names, axes ticks) set and randomly select and assign these labels to the entities. The values of the entities are floating-point numbers rounded to two decimal places, randomly selected from a uniform distribution ranging from 1 to 200.

\begin{figure*}
  \centering
  \includegraphics[width=1\linewidth]{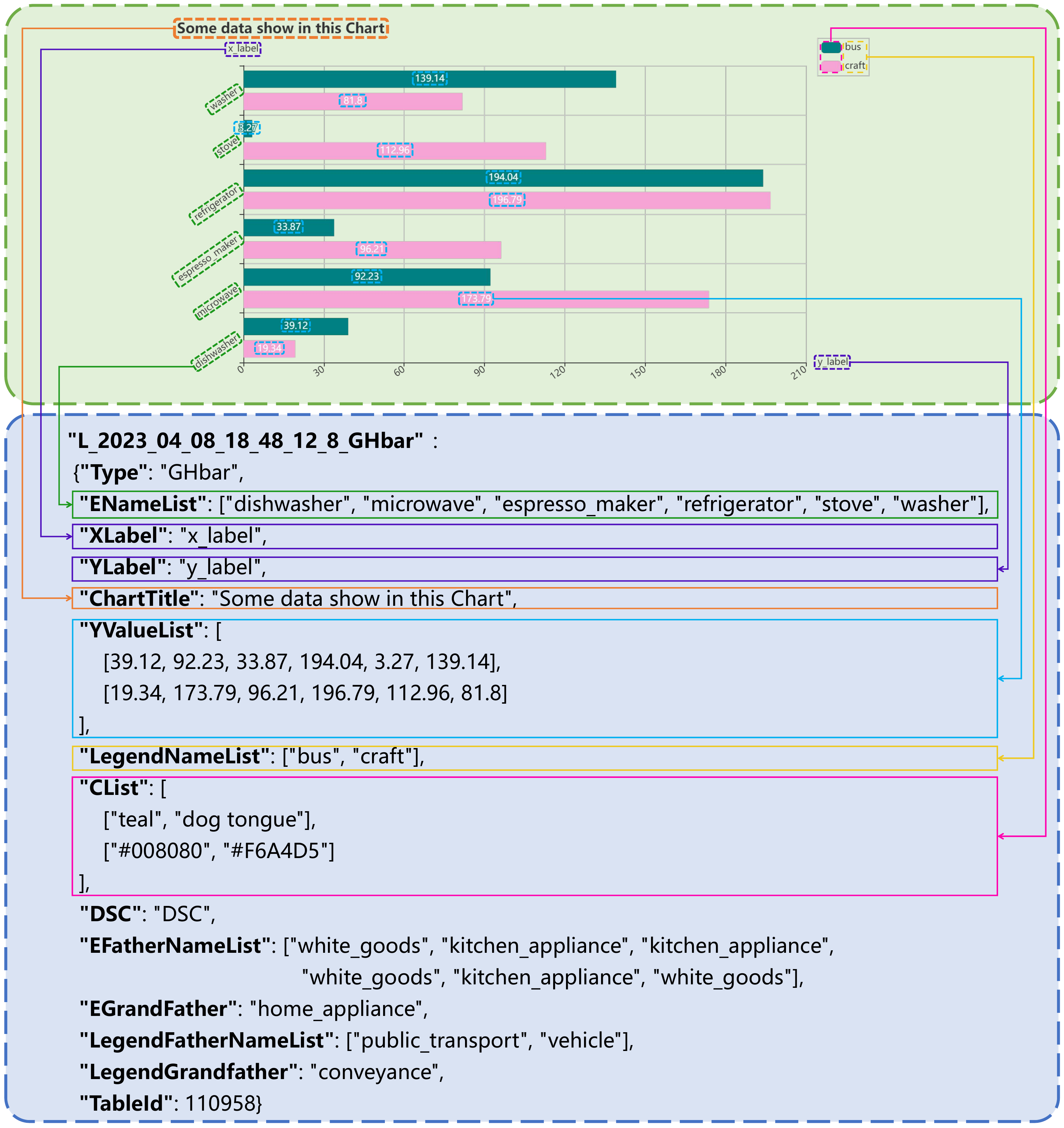}
  \caption{An example of chart infomation. The information includes the chart's title, entity names, legend labels, values, color names, and corresponding color values, etc.}
  \label{chart_infomation2} 
\end{figure*}

\vspace{6pt}
\noindent
\textbf{Chart Generation. } The initial step of the process involves chart generation, whereby predefined chart styles are used for specific chart types. These styles are pre-configured using Pyecharts and the corresponding source data tables are prepared for each chart. The source data table has a 20\% chance of being randomly generated and an 80\% chance of being sourced from real-world tabular data. The chart in Figure~\ref{data_generate1} are generated using randomly generated tabular data. The entity names of the table are extracted from the hierarchical entity library adopted in this paper. The specific extraction method is described in detail in Appendix \ref{section:d_1}. Upon the availability of the source data table and the predefined chart styles, the Pyecharts tool is employed to generate the chart and retrieve specific chart information, as demonstrated in Figure~\ref{chart_infomation2}, which presents the intricate details of the chart.

\begin{figure*}[t]
  \centering
  \includegraphics[width=.92\linewidth]{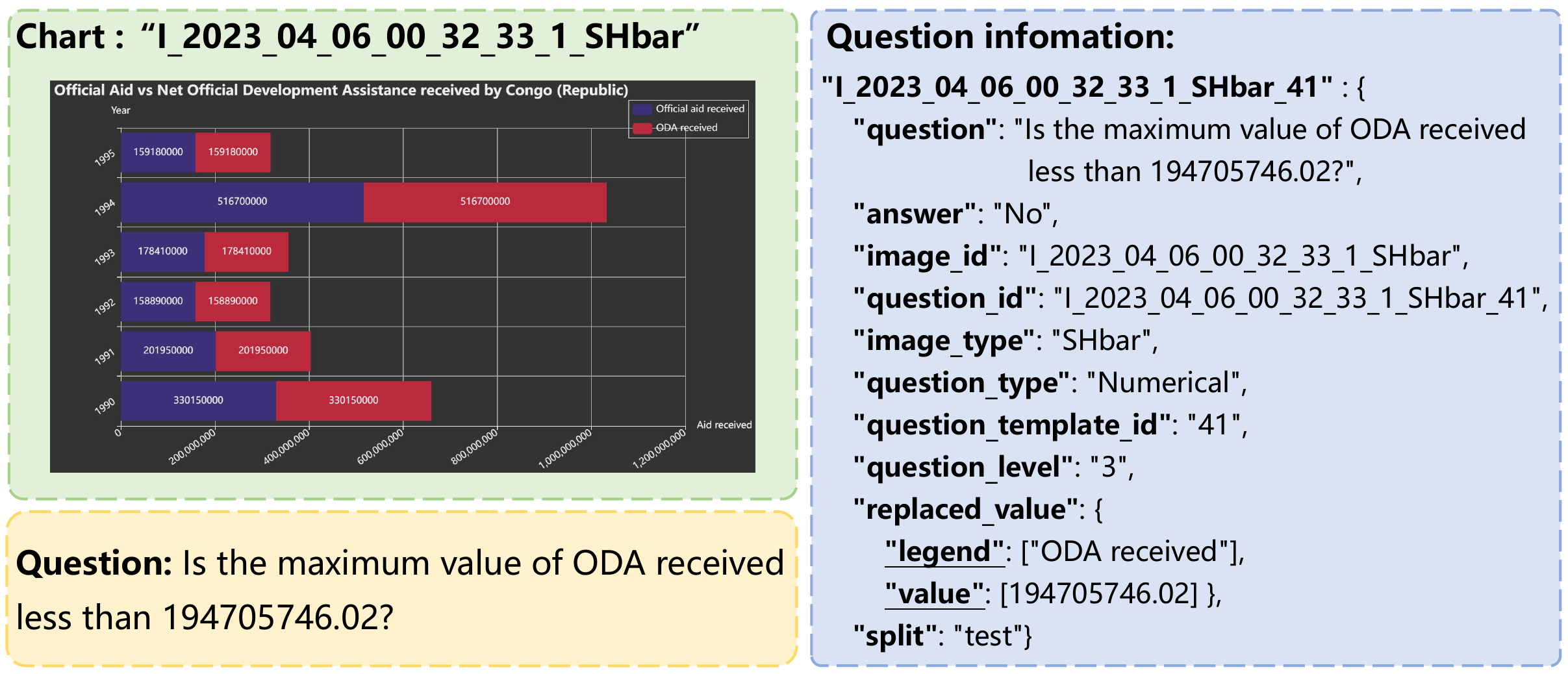}
  \caption{An example of question information. The information is composed of the question, answer, question ID, corresponding chart ID, template ID used for the question, difficulty level, and replaceable values, among other details.}
  \label{qinfo4}
\end{figure*}

\vspace{6pt}
\noindent
\textbf{Question Generation. } The question generation involves the preparation of question templates and chart information. As previously discussed, we distill and abstract the 5730 questions into 324 templates. Concretely, the template structure adheres to the following standard: \textit{"How many bars are greater than $<value>$ in the $<entity>$?"}. The placeholders represented by “\textit{$<entity>$}” are interchanged with the various graphical elements such as legends, X/Y ticks, the color of the legends/ticks and etc. The selection of values designated by “\textit{$<value>$}” is achieved via randomization that relates to the extent of range within the X/Y axis of the present chart. Based on the type of replaceable modules in the question template, chart-specific questions are generated by randomly selecting elements from the range of selectable replaceable module values in the chart information and filling them in the replaceable modules of the question template. The corresponding question information, as depicted in Figure~\ref{qinfo4}, can then be obtained. For instance, consider the template "What is the average of all bars between “\textit{$<entity\_1>$}” and “\textit{$<entity\_2>$}?" depicted in Figure~\ref{data_generate1}. The two replaceable modules, “\textit{$<entity\_1>$}” and “\textit{$<entity\_2>$}”, require substitution with specific entity names. To accomplish this, we randomly select two entity names from the entity name list provided in the chart information, namely "mouse" and "sunglass", respectively. Subsequently, we replace the placeholders with the chosen entities, resulting in the question "What is the average of all bars between mouse and sunglass?". The average length of the questions is 11.46 words.

\vspace{6pt}
\noindent
\textbf{Answer Generation. } The process of obtaining answers requires the provision of input parameters for solution derivation, which encompass the following parameters: (1) Image ID of the chart; (2) Question ID; and (3) Specific values to be filled into the replaceable modules within the question. The answer generation is achieved by designing solution steps and atomic operations. With this customized set of procedures, all answers for questions are created automatically without any human annotation, alleviating the probable errors and yielding cost-saving benefits. Moreover, we rigorously test the procedure for answer generation, with checks performed on each function and category of question, guaranteeing the correctness of the generated answer. The process of answer generation necessitates specific solution steps based on the question template, atomic operation functions that correspond to each step, filled-in replaceable values from the question information, and chart information. The atomic operation functions are invoked sequentially according to the question-solving steps, with the necessary parameters passed to the functions. These parameters are derived from the chart information and the replaced values of the substitutable modules for the question, resulting in the calculation of the answer. As an illustration in Figure ~\ref{data_generate1}, consider the question "What is the average value of all bars between mouse and sunglass?", according to the solving procedure, the first step involves calling the getEntityValue function with arguments 1 and 2, which indicate the retrieval of the replacement values for “\textit{$<entity\_1>$}” and “\textit{$<entity\_2>$}”, respectively corresponding to "mouse" and "sunglass". Subsequently, the getIntervalValueByEntity function is executed with "mouse" and "sunglass" as its parameters to retrieve the values of all bars located between these entities from the chart information, resulting in a list of values [85.6, 100.01, 101.1]. Ultimately, the function avg is invoked with the parameter [85.6, 100.01, 101.1] to compute the mean of all values in the list, yielding the answer to the query, which is 95.57.

\vspace{6pt}
\noindent
\textbf{Document Generation. } The document generation process entails the preparation of chart images, corresponding source data tables, and images from the COCO dataset. Initially, ChatGLM is employed to generate text content related to the table, which is then utilized as the primary textual information for the document. 
During the document generation process, based on LaTeX, a section of the document's y-axis is randomly selected as the next range to be filled. Then, either text, chart images, or images from the COCO dataset are randomly chosen and inserted into the selected section. The aforementioned process is repeated until the entirety of the y-axis range in the document is filled. For documents with two or three columns, the above process is executed sequentially for each column to complete the document generation.As the document is  progressively populated, the positional coordinates of each element within the document are systematically recorded and stored as annotation information within an XML document. This meticulous process allows us to obtain documents enriched with granular positional annotations. Further details regarding document information can be found in Appendix \ref{section:e}.

\section{Chart Infomation}
\label{section:c}
Chart information is stored in dictionary, where each chart's unique ID serves as the key. This allows for easy retrieval of chart information using its ID. An illustration of chart information is presented in Figure~\ref{chart_infomation2}.

Figure~\ref{chart_infomation2} shows the chart information of a grouped horizontal bar chart. 
The chart ID of this chart, “L\_2023\_04\_08\_18 \\ \_48\_12\_8\_GHbar” 
is composed of the “Machine ID\_ Generation time(Year)\_Generation time(Month) \_Generation time(Day)\_Generation time(Hour)\_Generation time(Minute) \\ \_Generation time(Sec-ond)\_Random(0-9)\_Chart type” used to generate the chart. The remaining content constitutes the chart information, including the chart type, title, entity name list (i.e., axis label), legend label list, data, element color name and color value, entity's parent class list, entity's grandparent class, legend label's parent class list, legend label's grandparent class, x-axis title, y-axis title, the table ID used to generate the chart, and some additional information (DSC). The additional information is provided for special chart types and is used to store some information not typically included in conventional charts, such as the value of marker lines in Marker Single Line and the range of each highlighted interval in Interval Highlight Single Line.

\section{Question Generation}
\label{section:d}
This section aims to provide a comprehensive introduction to two crucial components in the field of question generation, namely the hierarchical entity database and question information. 

\subsection{Hierarchical Entity Database}
\label{section:d_1}
The hierarchical entity database presented in this paper is a tree-like structure constructed from the labels found in ImageNet and WordNet, where the nodes with parent-child relationships represent hierarchical dependencies, i.e., a parent node includes child nodes and the parent node is the parent class of the child node. Specifically, the construction of a hierarchical entity database are as follows: (1) We opt for labels from ImageNet as fundamental entities, the hypernym query function in WordNet is employed to collect the hypernyms (i.e., parent nodes and ancestor nodes) of all the fundamental entities. (2) Entities with only a single child node among their hypernyms are excluded, and their parent and child nodes are connected $(A \rightarrow B \rightarrow C \Rightarrow A \rightarrow C)$ to eliminate redundant entities. (3) Finally, we obtain a directed acyclic graph (DAG), as some nodes may have multiple parent nodes. In this case, we retain only one parent node for these nodes and remove edges to other parent nodes. We save this entity repository as a tree structure in json file, which comprises noun words from WordNet that are utilized in the ImageNet dataset and their related words. All of the entities for generated data are randomly picked up from this hierarchical entity database. An example of the hierarchical entity database is provided in Appendix \ref{section:d_1}. When generating charts, multiple parent entities under the same grandparent class are selected, and multiple child entities are then extracted from each parent entity as entity names or legend labels. The entity names/legend labels and their corresponding parent and grandparent entities are recorded. When generating questions from randomly generated data, the replaceable modules in the templates are randomly selected from the parent entities for filling. For example, in the template \textit{"What is the sum of $<legend sort> in <entity sort>$?"}, if encountering $<entity sort>$, a parent entity name is randomly selected from the collection of parent entities of the entity name to fill in $<entity sort>$. If encountering $<legend sort>$, a parent entity name is randomly selected from the collection of parent entities of the legend label to fill in $<legend sort>$.

\begin{figure*}[t]
  \centering
  \includegraphics[width=0.88\linewidth]{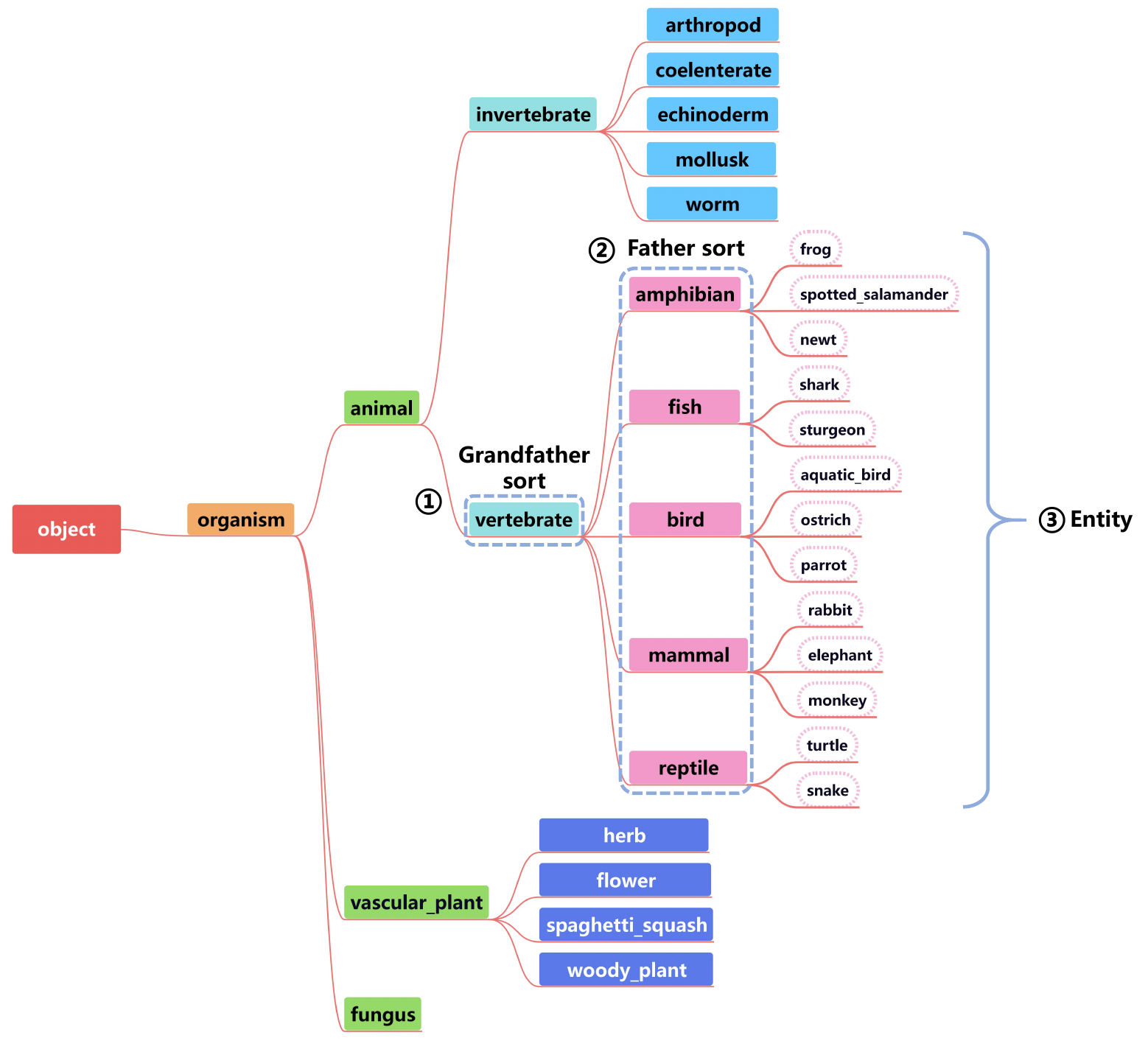}
  \caption{An example of our hierarchical entity library. This example is a part of the hierarchical entity library used in this paper. Vertebrates are selected as the grandparent node, and the children nodes under vertebrates are chosen as parent nodes. The resulting children nodes are used as entity names for the chart.}
  \label{hierarchical_entity_library3}
\end{figure*}

Figure~\ref{hierarchical_entity_library3} illustrates a part of the hierarchical entity database in this paper, where "organism" has three subclasses: "animal," "vascular\_plant," and "fungus," and each of these subclasses has its own subclasses. To extract the desired entities from the hierarchical entity database, a grandfather class is first selected, which in this case is "vertebrate" as shown in Figure~\ref{hierarchical_entity_library3}. Then, nodes are selected from the grandfather class's child nodes to serve as parent classes, and finally, the subclasses of the selected parent classes are identified as the required entities. The entities and their corresponding parent and grandfather classes are recorded in the chart information, as depicted in Figure~\ref{chart_infomation2}. Entities extracted from the hierarchical entity database can serve as entity names or legend labels in charts. They are extensively utilized in the chart generation, question formulation, and answer creation for charts generated from random data sources, and are present throughout the entirety of the dataset generation process.

\subsection{Question Information}
\label{section:d_2}
In a manner similar to the chart information, each piece of question information is systematically structured as a dictionary value, with a unique ID for each question serving as the key. This key-based organization enables the retrieval of information associated with any specific question by accessing its corresponding ID. Figure~\ref{qinfo4} illustrates an instance of question information along with its corresponding chart.

Figure ~\ref{qinfo4} displays an illustrative example of a stacked horizontal bar chart pertaining to an extremum question, with a question ID of "I\_2023\_04\_06\_00\_32\_33\_1\_S-Hbar\_41" generated by concatenating the chart ID and the question template ID. The dictionary value comprises comprehensive details about the question, including the question statement, answer, corresponding chart ID, chart type, question ID, question type, question template ID, question difficulty, fill-in values for replaceable modules in the question, and the dataset partition. These extensive details cater to the diverse requirements of researchers.

\section{Document Infomation}
\label{section:e}
\begin{figure*}
\centering
\includegraphics[width=1\linewidth]{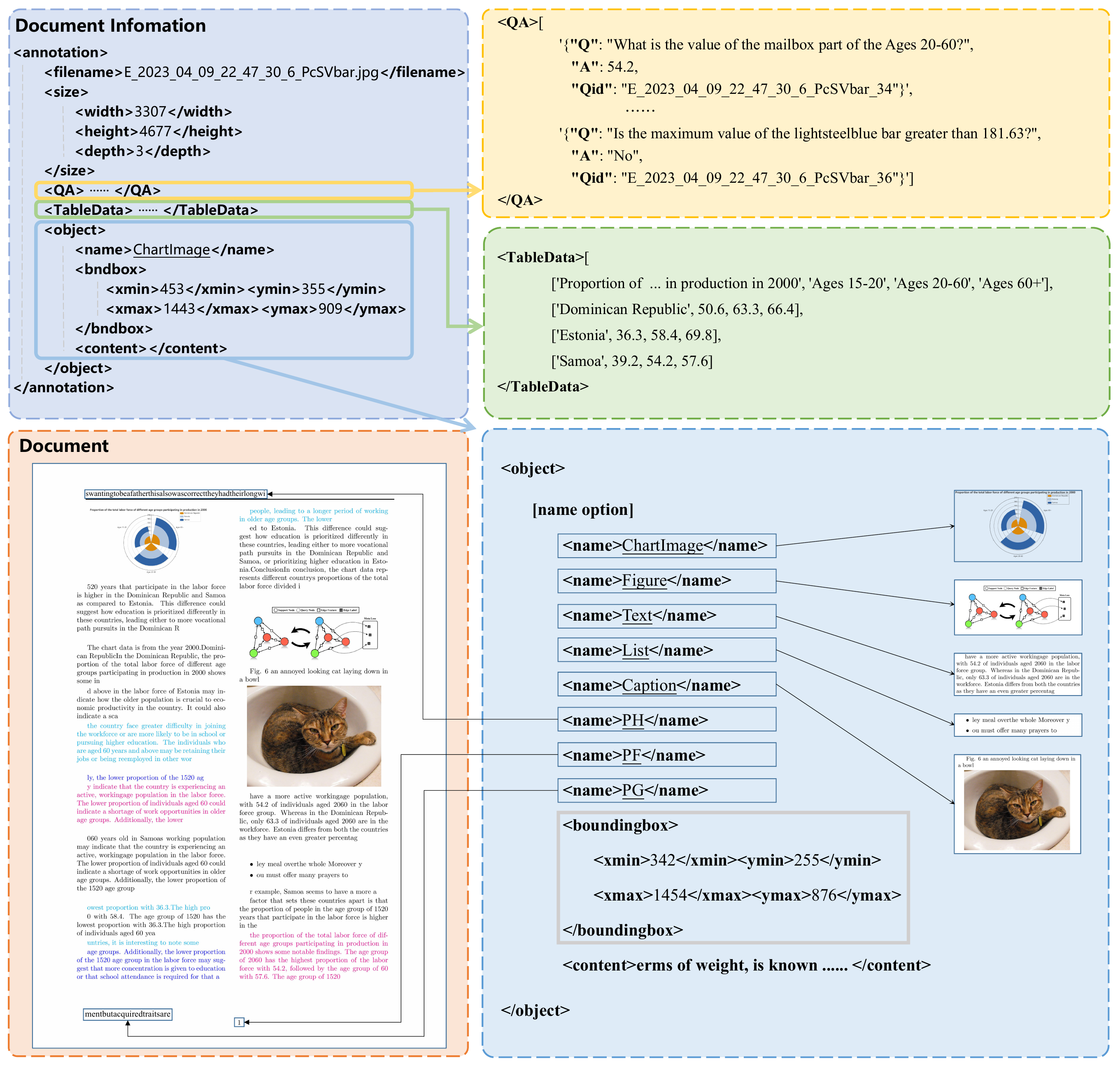}
\caption{An example of document information. The document image is situated in the lower-left corner, while a summarized version of the document information is located in the upper-left corner. For each highlighted section, detailed information is presented within a corresponding dashed box, labeled with the same color as the corresponding arrow.}
\label{document_infomation6}
\end{figure*}

The information within a document is stored in XML format, where each element's details are organized using tags. The elements within the document can be categorized into various types, such as chart images, regular images from the COCO dataset, text information pertaining to tables, textual content arranged as ordered or unordered lists, image descriptions, headers, footers, and page numbers. Each document element is encapsulated within an $<object>$ tag and comprises $<name>$, $<boundingbox>$, and $<content>$ components. The $<name>$ tag specifies the element's classification, while the $<boundingbox>$ indicates its position within the document. The position is determined by the pixel coordinates of the element's top-left and bottom-right corners, with the top-left corner of the document serving as the reference point. The <content> tag is utilized to store textual information associated with text elements, remaining empty for non-textual elements. In addition to the document elements, each document also retains the tabular data and all corresponding question-answer pairs related to the associated chart, as depicted in Figure~\ref{document_infomation6}.

\section{Document Examples}
\label{section:f}
\begin{figure*}[t]
  \centering
  \includegraphics[width=0.85\linewidth]{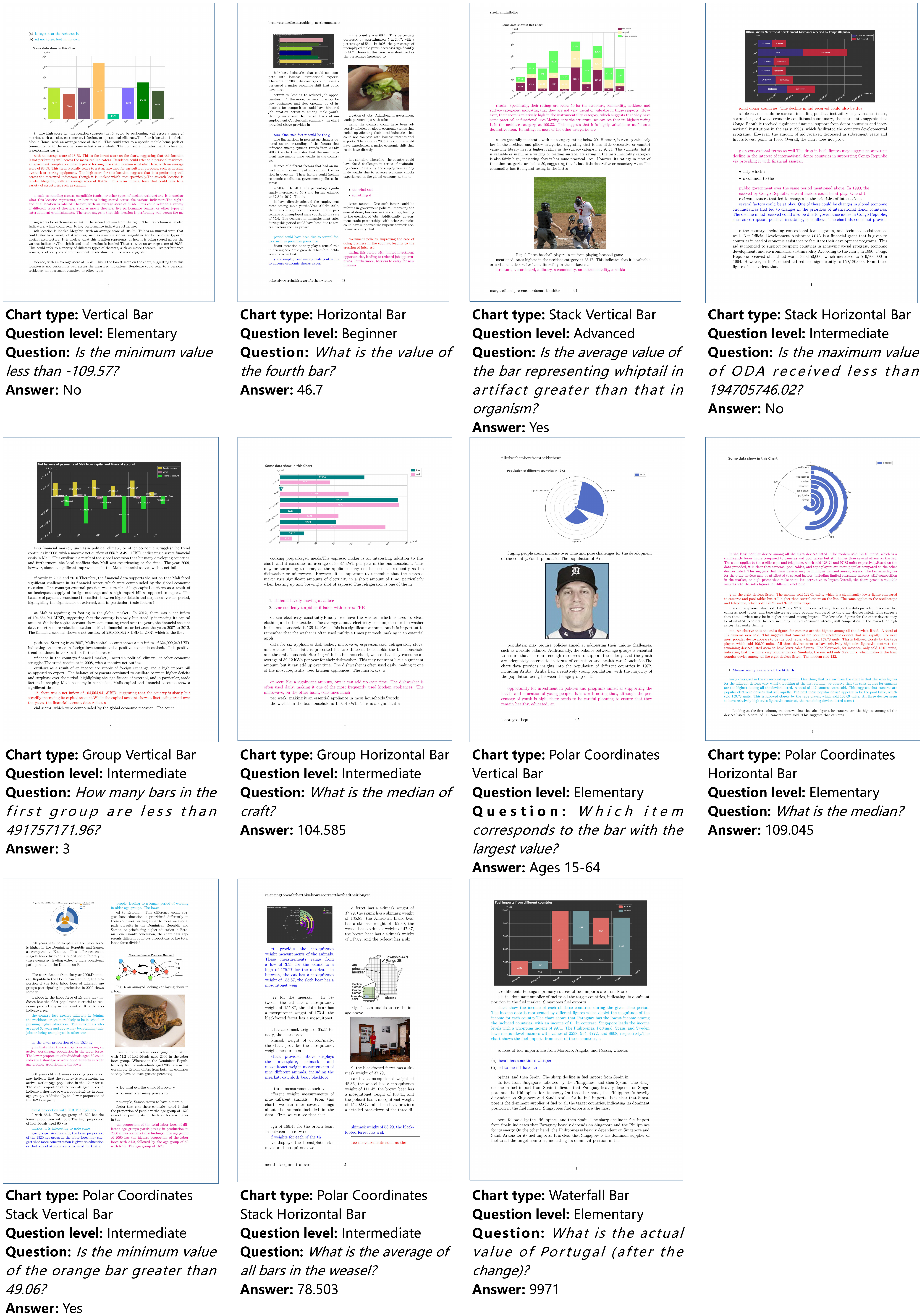}
  \caption{Examples of all subtypes of bar chart in the generated document. Each chart is accompanied by a question and corresponding 
answer.}
  \label{chart_fine-grained_sort_examples_bar}
\end{figure*}

\begin{figure*}[t]
  \centering
  \includegraphics[width=1\linewidth]{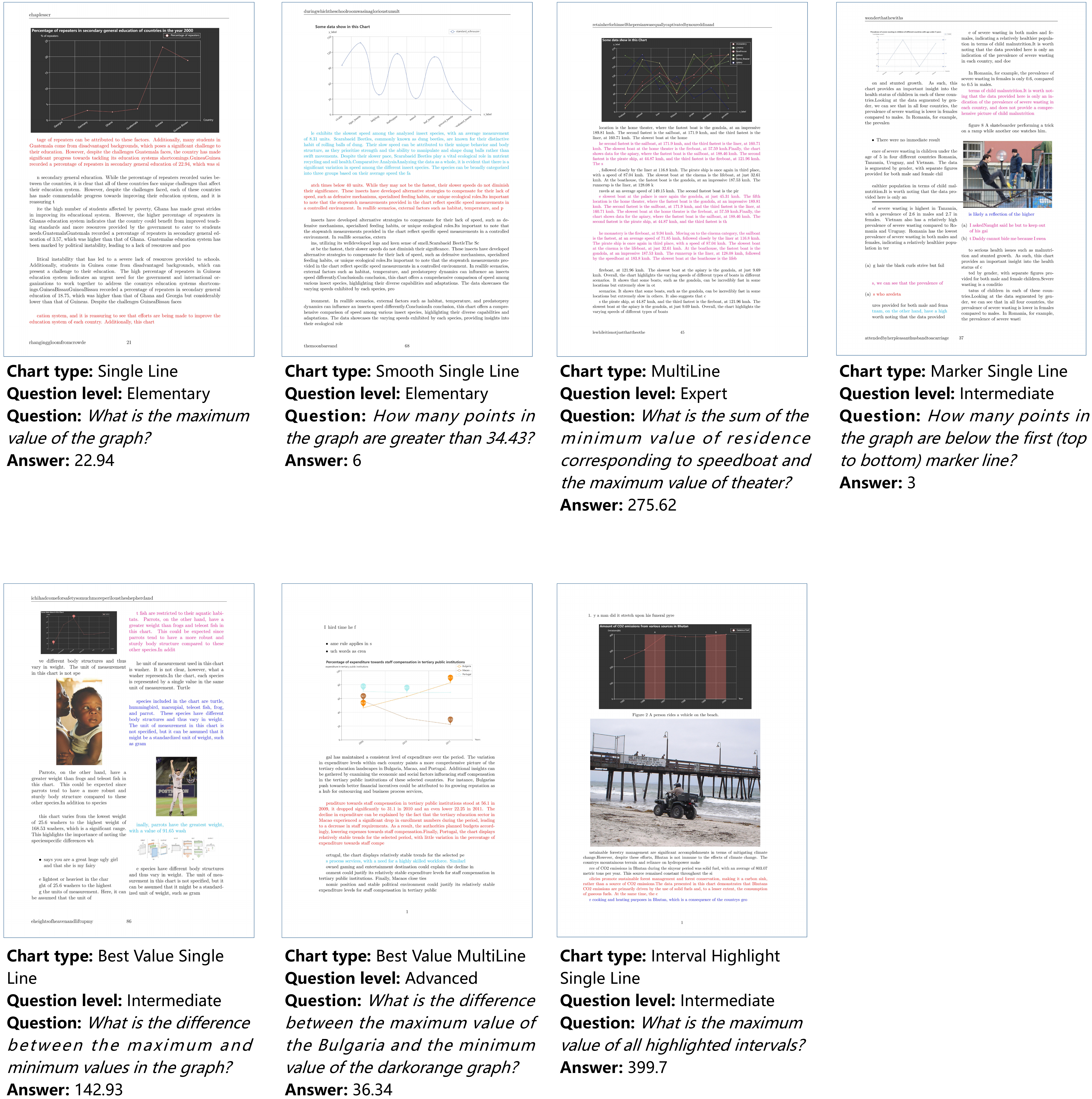}
  \caption{Examples of all subtypes of line plot in the generated document. Each chart is accompanied by a question and corresponding 
answer.}
  \label{chart_fine-grained_sort_examples_line}
\end{figure*}

\begin{figure*}[t]
  \centering
  \includegraphics[width=1\linewidth]{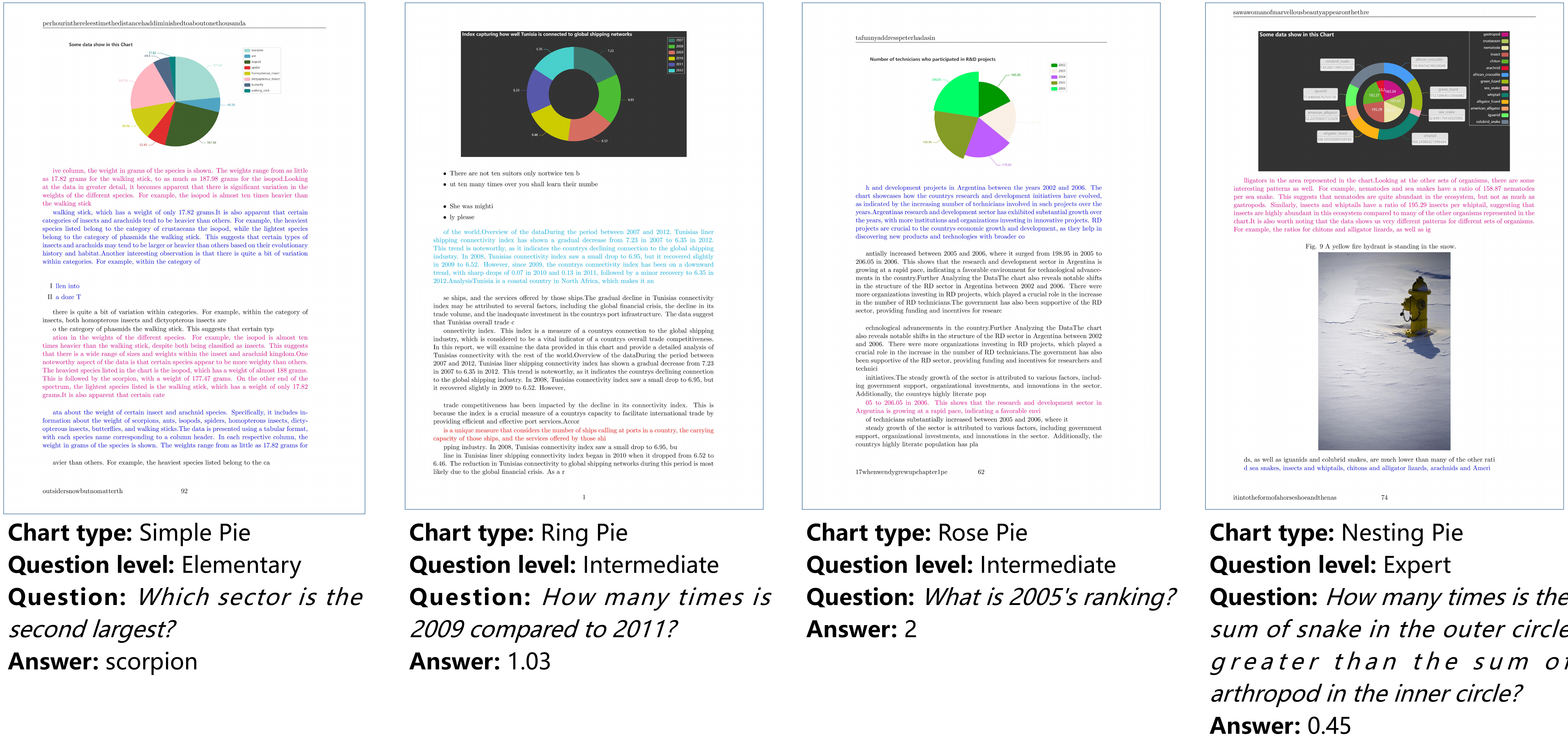}
  \caption{Examples of all subtypes of pie chart in the generated document. Each chart is accompanied by a question and corresponding 
answer.}
  \label{chart_fine-grained_sort_examples_pie}
\end{figure*}

\begin{figure*}[t]
  \centering
  \includegraphics[width=1\linewidth]{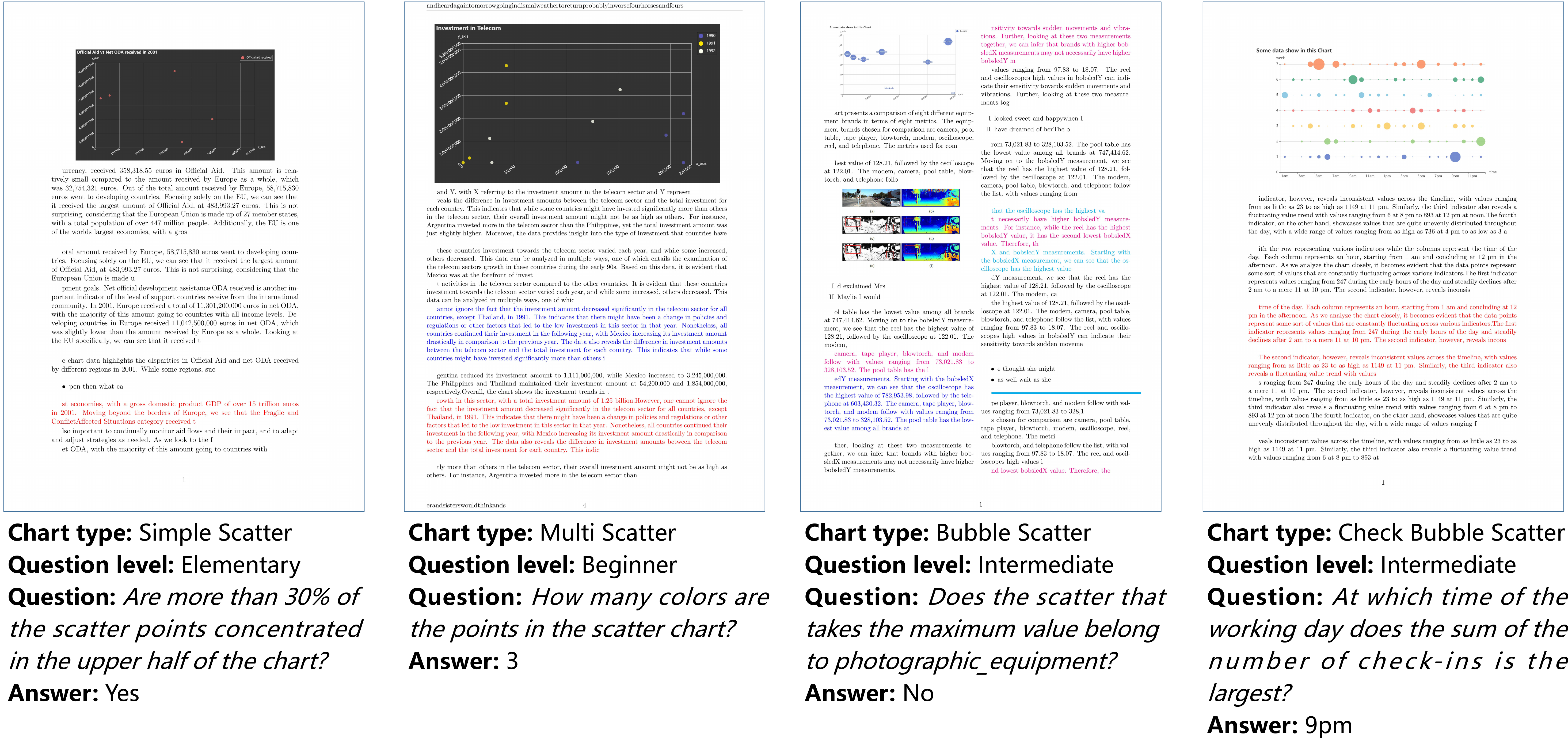}
  \caption{Examples of all subtypes of scatter plot in the generated document. Each chart is accompanied by a question and corresponding 
answer.}
  \label{chart_fine-grained_sort_examples_scatter}
\end{figure*}

\begin{figure*}[t]
  \centering
  \includegraphics[width=1\linewidth]{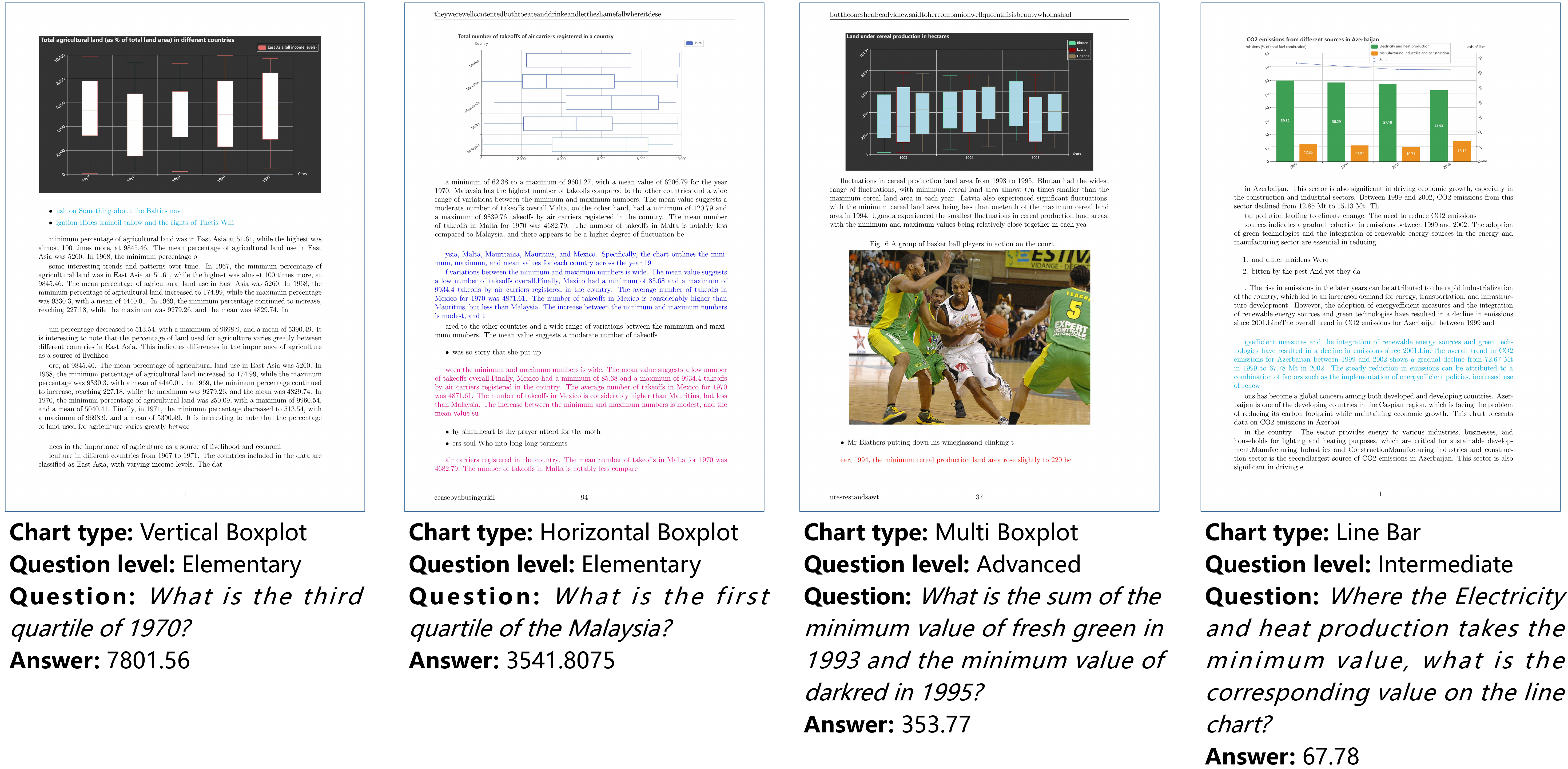}
  \caption{Examples of all subtypes of box plot and combination chart in the generated document. Each chart is accompanied by a question and corresponding 
answer.}
  \label{chart_fine-grained_sort_examples_box_and_comb}
\end{figure*}

The DCQA dataset encompasses a broad spectrum of chart types, encompassing 6 primary categories: bar chart, line chart, pie chart, scatter plot, box plot, and combination chart. Each category is further subdivided into a total of 30 subtypes. In this section, we will present a comprehensive enumeration of each subtype, accompanied by illustrative examples depicted in Figures~\ref{chart_fine-grained_sort_examples_bar} to ~\ref{chart_fine-grained_sort_examples_box_and_comb}.

\vspace{6pt}
\noindent
\textbf{Bar Chart.} Within the DCQA dataset, the bar chart category consists of 11 distinct subtypes: namely Vertical Bar, Horizontal Bar, Stack Vertical Bar, Stack Horizontal Bar, Group Vertical Bar, Group Horizontal Bar, Polar Coordinates Vertical Bar, Polar Coordinates Horizontal Bar, Polar Coordinates Stack Vertical Bar, Polar Coordinates Stack Horizontal Bar, and Waterfall Bar. Each subtype of the bar chart is exemplified in Figure ~\ref{chart_fine-grained_sort_examples_bar}, accompanied by a corresponding question-answer pair and an assigned difficulty level. The objective of this presentation is to improve understanding and perception of the dataset's style and level of difficulty.

\vspace{6pt}
\noindent
\textbf{Line Chart.} The line chart category primarily encompasses 7 subtypes, namely Single Line, Smooth Single Line, MultiLine, Marker Single Line, Best Value Single Line, Best Value MultiLine, and Interval Highlight Single Line. Each subtype of the line chart is exemplified by an example depicted in Figure ~\ref{chart_fine-grained_sort_examples_line}.

\vspace{6pt}
\noindent
\textbf{Pie Chart.} Pie chart comprise 4 main subtypes, specifically Simple Pie, Ring Pie, Rose Pie, and Nesting Pie. Examples of each pie chart subtype is provided in Figure ~\ref{chart_fine-grained_sort_examples_pie}.

\vspace{6pt}
\noindent
\textbf{Scatter Plot.} Scatter plot comprise 4 main subtypes, namely Simple Scatter, Multi Scatter, Bubble Scatter, and Check Bubble Scatter. An example of each subtype is illustrated in Figure ~\ref{chart_fine-grained_sort_examples_scatter}.

\vspace{6pt}
\noindent
\textbf{Box Plot and Combination Chart.} Box plot encompass 3 distinct subtypes, namely Vertical Boxplot, Horizontal Boxplot, and Multi Boxplot. In contrast, combination chart comprise a single subtype, specifically Line Bar. An illustrative example for each subtype of the box plot and combination chart is presented in Figure ~\ref{chart_fine-grained_sort_examples_box_and_comb}.

\end{document}